\documentclass{article}

\PassOptionsToPackage{numbers, compress}{natbib}
\usepackage[preprint]{neurips_2024}

\usepackage[utf8]{inputenc} 
\usepackage[T1]{fontenc}    
\usepackage{hyperref}       
\usepackage{url}            
\usepackage{booktabs}       
\usepackage{amsfonts}       
\usepackage{nicefrac}       
\usepackage{microtype}      
\usepackage{xcolor}         
\usepackage{graphicx}
\usepackage{amsmath} 
\usepackage{amsthm}
\usepackage{amssymb}
\usepackage{adjustbox}
\usepackage[capitalize]{cleveref}
\usepackage{graphicx}
\usepackage{subcaption}
\usepackage{caption}

\newtheorem{theorem}{Theorem}[section]

\newtheorem{lemma}[theorem]{Lemma}
\newtheorem*{remark}{Remark}
\theoremstyle{definition}

\usepackage{amsthm}

\newcounter{subassumption}[asu]

\makeatletter
\renewcommand{\p@subassumption}{\theasu}
\makeatother

\title{Feature Mapping in Physics-Informed Neural Networks (PINNs)}

\author{Chengxi Zeng\thanks{Correspondence Author. \\
The code can be found in repo \url{github.com/SimonZeng7108/RBF-PINN/tree/master}.}, Tilo Burghardt \& Alberto M.~Gambaruto  \\
University of Bristol, UK\\
\texttt{\{cz15306, tb2935, alberto.gambaruto\}@bristol.ac.uk} \\
}

\begin{document}

\maketitle
\vspace{-0.2in}
\begin{abstract}
\vspace{-0.1in}
In this paper, the training dynamics of PINNs with a feature mapping layer via the limiting Conjugate Kernel and Neural Tangent Kernel is investigated, shedding light on the convergence of PINNs; Although the commonly used Fourier-based feature mapping has achieved great success, we show its inadequacy in some physics scenarios. Via these two scopes, we propose conditionally positive definite Radial Basis Function as a better alternative. Lastly, we explore the feature mapping numerically in wide neural networks. Our empirical results reveal the efficacy of our method in diverse forward and inverse problem sets. Composing feature functions is found to be a practical  way to address the expressivity and generalisability trade-off, viz., tuning the bandwidth of the kernels and the surjectivity of the feature mapping function. This simple technique can be implemented for coordinate inputs and benefits the broader PINNs research.
\end{abstract}

\vspace{-0.2in}
\section{Introduction}
\vspace{-0.1in}
\label{Introduction}
Our observed world is described by the laws of physics, and many phenomena can be defined by sets of Differential Equations (DEs). The learning paradigm that enforces the mathematical rules and makes use of the available data is called Physics-Informed Machine Learning (PIML)~\cite{Karniadakis2021PhysicsinformedML}. Physics-Driven approaches have recently achieved significant success in a wide range of leading scientific research, from Electronics~\cite{Smith2022PhysicsInformedIR, Hu2023SyncTREE, Nicoli2023Quant} and Medical Image~\cite{Goyeneche2023ResoNet, Salehi2021PhysGNNAP, pokkunuru2023improved} to Dynamical System~\cite{Thangamuthu2022Dynamical, Ni2022NTFieldsNT} and Meteorology~\cite{kashinath2021physics, Giladi2021PhysicsAwareDW}. Among these, one of the most prominent methods is termed Physics-Informed Neural Networks (PINNs)~\cite{Raissi2019PhysicsinformedNN}. It leverages the expressivity and differentiability of deep Neural Networks (NN) and integrates the DEs in the NN as a regulariser to introduce strong inductive biases during training. PINNs are also considered as a special type of Neural Fields~\cite{Xie2021NeuralFI}.

PINNs share many common challenges, faltering at accurate convergence that is referred to as `failure modes' of PINNs. \citet{Wang2020WhenAW} leverage the `Neural Tangent Kernel' theory that reveals PINNs suffer from `Spectral Bias' due to the `lazy-training' regime~\cite{Geiger2019DisentanglingFA}; \citet{Krishnapriyan2021CharacterizingPF} demonstrate that PINNs are inherently difficult to optimise in harder Partial Differential Equations (PDEs) and multi-dimensional space; Lack of symmetry in the distribution of PDE and imbalanced residuals resulting in solutions from IC/BC cannot effectively alleviate the trivial solution in the PDE, which is described as `propagation failure' in~\cite{Daw22Propagation}. These analyses are principled in the design of PINNs variants, such as loss-reweighting~\cite{Wang2021UnderstandingAM, WANG2021113938, Wang2022RespectingCI, Psaros2021MetalearningPL}, domain decomposition~\cite{Jagtap2020ExtendedPN, Kharazmi2020hpVPINNsVP, Moseley2021FiniteBP, Li2022MetaLO}, stronger regularisation~\cite{Yu2022Gradient, Wang22l2, AkhoundSadegh2023LiePS} amongst others. 

Whilst notable progress has been made in previous work, feature mapping has not been thoroughly studied, with only few work~\cite{WANG2021113938,Wong2022Sinusoidal} finding its potential in PINNs. Feature mapping was initially proposed in Natural Language Processing (NLP) with the goal to map the input to a high-dimensional feature space. It was later found effective at tackling spectral bias in visual tasks~\cite{Tancik2020FourierFL}.
\begin{figure}[ht]
\setlength{\belowcaptionskip}{-30pt}
\begin{center}
\centerline{\includegraphics[height=3.5cm]{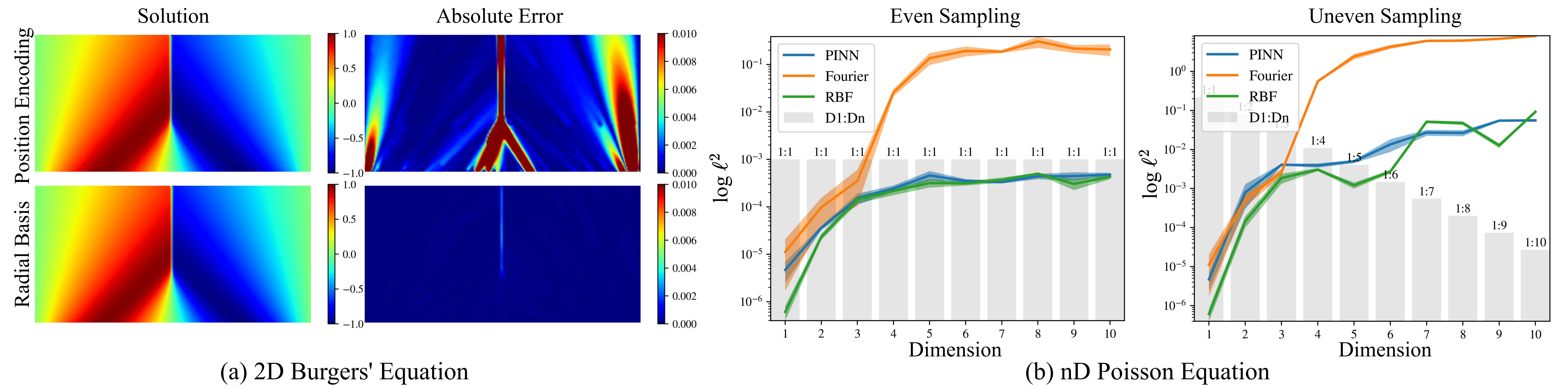}}
\vspace{-0.1in}
\caption{(a) Fourier based Positional Encoding shows inadequate generalisation at the discontinuity in Burgers' Equation; (b) Random Fourier Features fail at high dimensional Poisson Equation. Error on nD Poisson equation from 1 to 10 dimensions cases (left), and a more realistic setting with uneven sampling on each dimension (right). The experiments are repeated 3 times with different random seeds, and the variances are highlighted in shades.}
\label{Fig: error demo}
\end{center}
\end{figure}

In this study, we are motivated by the potency of feature mapping in the wider neural representation research.  In the infinite-width limit, we establish theoretical study of the training dynamics of PINNs with a feature mapping layer through the lens of linking two kernels: the Conjugate Kernel (CK) and Neural Tangent Kernel (NTK). The CK, which directly links to feature mapping, is largely overlooked in the PINNs community. Specifically, the limiting CKs are sensitive to the inputs and network parameters initialisation. Moreover, they depend on the input gradient of the model, the variance of each layer and the non-linear activation functions that they pass through~\cite{Hu2021OnTR}. Hence, the convergence of the PINNs are strongly influenced by the features before the parameterised layers.

There are two main characteristics imposed by the CK and NTK in the infinite-width limit. Firstly, in this regime, the neural network behaves as a linear model, which can be analysed in a conventional regression setting. With an appropriate initialisation and loss functions, the training loss converges to zero. Moreover, gradient descent is able to find the global minimum with unchanged parameters. This is confirmed in two-layer PINNs by~\cite{Gao2023GradientDF}. Secondly, the spectra from the decompositions of the CK and NTK are elongated by random initialisations~\cite{Fan2020SpectraOT}. The associated eigenvectors are the main factors driving the training dynamics of the neural network, suggesting that they govern the generalisation property in an overparameterised model. 

We show that the coordinate-based input after a feature mapping layer positively impacts the CK and NTK. As a result, it improves the overall convergence of the model training. Subsequently, we propose a framework for the design of the feature map layer that helps CK and NTK propagate in a practical setting. 
Our contribution can be summarised as follows:
\vspace{-0.1in}
\begin{itemize}
 \item {We provide theoretical work on the training dynamics of PINNs with a feature mapping layer in the limiting Conjugate Kernel and Neural Tangent Kernel scope (\Cref{Prop: ck} and \Cref{theorem: ntk}). It reveals that the initial distribution of the feature mapping layer determines the propagation of the two kernels and the important properties of the mapping function.}
 \item {We show the limitations and failures of the common Fourier-based feature mapping in some Partial Differential Equations and justify such mathematical behaviour by its cardinality, i.e., Fourier functions are highly surjective (\Cref{lemma: inject}).}
 \item {We study feature mapping in practical settings and demonstrate a general framework for the design of feature mapping and propose conditional positive definite Radial Basis Function, which outperforms Fourier-based feature mapping in a range of forward and inverse tasks.}
\end{itemize}

\vspace{-0.2in}
\section{Background and Prior theories}
\vspace{-0.1in}
\subsection{Physics-Informed Neural Network}
\vspace{-0.1in}
Conforming to traditional solvers, the formulation of PINNs requires initial/boundary conditions (IC/BC) in a bounded spatial-temporal domain. The sampled points and any prescribed conditions (e.g., real-valued Dirichlet boundary condition) are trained along with the collocation points that evaluate the residuals of the DEs. The goal is to optimise the overparameterised NN by minimising the residuals. Such converged parameter space can hence constitute a surrogate model that represents the solution space of the DEs. Following the formulation by \citet{Raissi2019PhysicsinformedNN}, the Physics-Informed Neural Network that solves both forward and inverse problems in PDEs is reviewed in a general form:
\begin{equation}
\left\{ \begin{aligned} 
\mathcal{D}[u(\mathbf{x}, t; \alpha_i)] &= F(\mathbf{x}, t) & \quad t \in \mathcal{T}[0, T], \forall \mathbf{x} \in \Omega,\\
u(\mathbf{x}, 0) &= G(\mathbf{x}) & \quad \mathbf{x} \in \Omega,\\
\mathcal{B}[u(\mathbf{x}, t)] &= H(\mathbf{x}, t) & \quad t \in \mathcal{T}[0, T], \mathbf{x} \in \partial\Omega,
\end{aligned} \right.
\end{equation}
\vskip -0.1in
where $\mathcal{D}[\cdot]$ is the differential operator, $\mathbf{x}$ and $t$ are the independent variables in spatial and temporal domains $\Omega$ and $\mathcal{T}$, respectively. The $\alpha_i$ are coefficients of the DE system and remain wholly or partially unknown in inverse problems. The DE confines to the initial condition when $t=0$ and the boundary operator $\mathcal{B}$ at the boundary $\partial \Omega$. $F$, $G$ and $H$ are arbitrary functions.

PINNs are parameterised by $\theta$ that can solve $\hat u_\theta$ at any $\mathbf{x}$ and $t$ in the domain, hence the training loss functions are defined as follows:
\setlength{\abovedisplayskip}{0pt}
\begin{equation}
\label{equation: loss}
\begin{split}
\mathcal{L}(\theta; \mathbf{x}_{(x, t)})  = \frac{\lambda_r}{N_r}\sum_{i=1}^{N_r} \left | \mathcal{D}[\hat u_\theta(x_{r}^i)] - F(x_r^i) \right |^2 & + \frac{\lambda_{ic}}{N_{ic}}\sum_{i=1}^{N_{ic}} \left | \hat u_\theta(x_{ic}^i) - G(x_{ic}^i) \right |^2 \\  & + 
\frac{\lambda_{bc}}{N_{bc}}\sum_{i=1}^{N_{bc}} \left | \mathcal{B}[\hat u_\theta(x_{bc}^i)] - H(x_{bc}^i) \right |^2,
\end{split}
\end{equation}
\vskip -0.15in
where $\{x_r^i\}_{i = 1}^{N_r}$, $\{x_{ic}^i\}_{i = 1}^{N_{ic}}$ and $\{x_{bc}^i\}_{i = 1}^{N_{bc}}$ are collocation points, initial condition points and boundary condition points sampled from the bounded domain and evaluated by computing the mean squared error. $\lambda_r$, $\lambda_{ic}$ and $\lambda_{bc}$ are the corresponding weights of each term. In this paper, we consider solving forward DE problem in an unsupervised learning setting, though additional experimental/data points can be simply added to form the loss term $\mathcal{L}_{data}(\theta; x_{data})$ as a strong regulariser. 

\vspace{-0.1in}
\subsection{Feature Mapping in PINNs}
\vspace{-0.1in}
In the original form of PINNs, standard multi-layer perceptions (MLPs) have been adopted as the implicit neural presentation of the DEs. The MLPs of $L$ layers can be mathematically expressed in the following recursive formulation, where the model is parameterised by $\theta=\{\mathbf{W}^l, \mathbf{b}^l\}_{l = 1}^{L}$, $\mathbf{W}^l$ is the weight matrix of the $l$-th layer, and $\mathbf{b}^l$ is the trainable bias before the non-linear activation function $a$. At initialisation, each element of $\mathbf{W}$ and $\mathbf{b}$ are sampled independently from $\mathcal{N}(0, 1)$: 
\setlength{\abovedisplayskip}{0pt}
\begin{equation}
\label{Equa: MLP}
\begin{aligned}
f_i^{l}(\mathbf{x}; \theta) &= (b^{l}_i + \frac{1}{\sqrt{d^l}}\sum_{j}w^{l}_{ij} h_{j}^{l-1}(\mathbf{x}; \theta) ); \quad  
h_i^{l}(\mathbf{x}; \theta) = a \left(f_i^{l}(\mathbf{x}; \theta) \right) ,
\end{aligned}
\end{equation}
\vskip -0.15in
where $h_i^l$ is the output of the $i$-th neuron at the $l$-th layer.
The normalisation $\frac{1}{\sqrt{d^l}}$ of weights by width $d^l$ is placed so that we can take the width of the layers to infinity in the wide neural network regime. Feature mapping in the first layer is defined as:
\setlength{\abovedisplayskip}{0pt}
\begin{equation}
\label{Equa: feature mapping}
f_i^{1} =\frac{1}{\sqrt{d^1}}\sum_{j}w^{1}_{ij}\varphi_j(\mathbf{x}; \theta) \ ,
\end{equation}
\vskip -0.15in
where $x$ is the coordinate-based input, $\varphi$ is a feature mapping operator that projects input to a higher dimension feature space, $\Phi: x \in \mathbb{R}^n \rightarrow \mathbb{R}^m$, and typically $n \ll m$. 

Feature mapping is a broader term for positional encoding that can involve either fixed encoding or trainable embedding. Examples of Fourier feature mappings and other methods are given in Appendix \ref{Append: FM}. Following, we introduce two theoretical works regarding feature mapping in PINNs.
\vspace{-0.1in}
\subsection{Prior Theories}
\vspace{-0.1in}
\textbf{Spectral Bias:} One work theoretically supports feature mapping in PINNs is given by~\cite{Wang2020WhenAW}, which formulates the training dynamics of PINNs following the seminal work in general MLPs~\cite{Jacot2018NeuralTK} and proves the PINNs model converges to a deterministic kernel when the width of the NN tends to infinity. As a result, the training of PINNs is dominated by the leading eigenvalues in the Neural Tangent Kernel (NTK), this is termed Spectral Bias (more details about Spectral Bias in PINNs can be found in Appendix~\ref{Append: Spectral}). Hence, a tunable bandwidth kernel is desirable to mitigate the Spectral Bias phenomenon. Bochner's theorem is employed to approximate a shift-invariant kernel with a controllable kernel width (example in Appendix~\ref{Append: CNTK}). It tends to be  useful to compute Fourier features in multi-scale or high-frequency physical cases, and the optimal value of $\sigma$ in each case can be identified by line search.

\textbf{Input Gradient Invariability:} Another proposition by~\cite{Wong2022Sinusoidal} suggests that PINNs suffer from limited input gradient variability under certain weight initialisation, which prevents the optimisation of parameter space from reaching joint PDE and BC optimal solutions. The key finding  they reveal is that it is the enhanced input gradient distribution that improves the performance (details in Appendix~\ref{Append: Input Gradient Variability}), not the features themselves. They employed the learnable Sinusoidal feature from concurrent work~\cite{Sitzmann2020ImplicitNR} which can increase input
gradient variability and help gradient descent escape the local minimum at initialisation.

In summary, the prior theories reveal the feature mapping is in favour of mitigating the spectral bias and increase input gradient variability. However, a complete theory addressing how feature mapping particularly controls the training dynamics of PINNs with a feature mapping layer and the principle of design a feature mapping layer is still missing. Moreover, either of them showed the implementation limitations of the Fourier feature based methods. In the next subsection, we show two examples that Fourier-based features exemplify deficient functionality.
\vspace{-0.1in}
\subsection{Limitation of the Fourier Features}
\vspace{-0.1in}
Simple Fourier features can lead to undesirable artefacts, shown in Figure~\ref{Fig: error demo}~(a) (detailed equations are in Appendix~\ref{Append: Benchmark}). When using Positional Encoding to solve Burgers' equations, there appears to be a high prediction error in the region approaching a discontinuous solution. This effect is analogous to the Gibbs phenomenon, that is the approximated function value by a finite number of terms in its Fourier series tends to overshoot and oscillate around a discontinuity. Such inferior interpolation performance of the Positional Encoding has also been observed in visual computing~\cite{Ramasinghe2021ALR}, it has shown the smoothness of the distortion of the manifold formed by the feature mapping layer is the key to memorisation and generalisation trade-off.

Another surprising experimental result which exhibits poor performance is using Random Fourier Features~\cite{Tancik2020FourierFL} in high-dimension problems, as demonstrated in Figure~\ref{Fig: error demo}~(b). An nD Poisson equation is tested from 1 to 10 dimensions with a Dirichlet boundary condition. Firstly, ten test cases with increasing dimensions are set up with a fixed number of collocation points and they are evenly sampled for each dimension (i.e., the ratio between each dimension is equal in each case, we denote the ratio between the first dimension to the last dimension by the bar chart). The total $\ell^2$ error increases when the dimension of each case increases, as it becomes harder to solve. An evident result is that the random Fourier feature mapping does not generalise well in high dimensions ($D>3$). Due to the global and smooth solution of the Poisson Equation and homogeneity across dimensions, the standard PINNs seem to perform on a par with an additional RBF layer in the even sampling case.

Secondly, a more realistic case is set up when the number of sampling points is not the same in different dimensions (Figure~\ref{Fig: error demo}~(b)~right). In this setting, we set the number of sampling points $x_r = \frac{1}{D}$ for each case, meaning that as the dimension increases, there are fewer sampled points. This resembles the training setting in the unsteady Navier-Stokes equations for fluids dynamics, which has fine sampling density in the spatial domain, but potentially rather sparse sampling in the temporal dimension. Although all methods have shown an increase in error, the PINNs with Random Fourier Features in particular has been significantly underperforming in higher dimensions. We tuned a few hyperparameters in the Fourier feature including the arbitrary scale $\sigma$ and the number of Fourier features, yet none of the trials reduced the high error in high dimension scenarios. 

To encapsulate the two observations, we suggest a new perspective on the matter and show in the following proof that the Fourier modes as a mapping function is likely to produce overlapping values.
\begin{lemma}
\label{lemma: inject}
Consider a randomly sampled and normalised input $\mathbf{x}=[x_1, x_2, \cdots, x_n]^T$, $x \in [0, 1]^d$, and its corresponding features in $\Phi: \mathbb{R}^d \rightarrow \mathbb{R}^m = [\varphi(x_1), \varphi(x_2), \cdots, \varphi(x_n)]^T$, let the feature mapping function $\varphi(\mathbf{x}) = sin(2\pi \mathcal{B} \mathbf{x}) \in [-1, 1]$, where $\mathcal{B}$ is sampled from a Gaussian distribution $\mathcal{N}(0, \sigma)$, the mapping function $sin(\cdot)$ is surjective w.h.p.
\end{lemma}
\vspace{-0.2in}
\begin{proof} 
see Appendix \ref{Append: inject}.
\end{proof}
\vspace{-0.1in}
A surjective function denotes that the inputs are redundantly mapped to the feature space from the domain. This indicates that an overlapping image is likely to be formed in the projected codomain, and it can also partially explain the Gibbs phenomenon in discontinuous regions with overshoot function values. As it can be easily inferred, when the input dimension gets higher, the probability of the Fourier features is even higher to be surjective. From this viewpoint, we provided additional theoretical support to~\cite{Wong2022Sinusoidal} regarding the generalisation of the PINNs, that is the function surjectivity limits input gradient variability and ultimately traps the parameter space to local minima.

\vspace{-0.15in}
\section{Training dynamics of PINNs}
\vspace{-0.1in}
\subsection{Theory Settings}
\vspace{-0.1in}
Given the training dataset $\mathbf{x} = \{x_r^i\}_{i = 1}^{N_r} \cup \{x_{ic}^i\}_{i = 1}^{N_{ic}} \cup \{x_{bc}^i\}_{i = 1}^{N_{bc}}$. 
In the infinite-width limits, the non-linear neural network evolves similarly to the kernel regression models~\cite{Lee2019WideNN}. Hence we can leverage two types of kernels, the \textbf{Conjugate Kernel (CK)} and the \textbf{Neural Tangent Kernel (NTK)} to analyse the initial distribution of the model and the training dynamics of the PINNs to the infinite-width limit.\\
The CK in each layer is defined as:
\begin{equation}
K_{CK}^l = {\mathcal{X}^l}^T\mathcal{X}^l\in \mathbb{R}^{N\times N} = \Sigma^l(\mathbf{x}, \mathbf{x}'),
\end{equation}
and we formulate the general (in contrast to the formulation in~\cite{Wang2020WhenAW}) NTK in PINNs as:
\begin{equation}
K_{NTK}^l = \nabla_{\theta}f^l(\mathcal{X}; \theta)^T\nabla_{\theta}f^l(\mathcal{X}; \theta)   \in \mathbb{R^{N\times N}} = \Theta^l(\mathbf{x}, \mathbf{x}'),
\end{equation}
Where $\mathcal{X}=\{\mathcal{X}_{\Phi}\}\bigvee \{\mathcal{X}_h\}$, $\mathcal{X}_{\Phi} = \varphi(\mathbf{x})\in\mathbb{R}^{N \times n} $ is the matrix after the feature mapping and $\mathcal{X}^{l}_{h} = h^l(\mathbf{x}) =\frac{1}{d^l} a(W^l \mathbf{x}^{l-1}) \in\mathbb{R}^{N \times d^l}$ is the matrix after the activation function.

\textbf{Assumptions.} (1) The number of layers in the PINNs, $L\geq 2$. The first layer is the feature mapping layer, and the layers after are the parameterised layers; (2) $d^1, \cdots, d^L \to \infty$;
(3) The weights in the layers are initialised by Xavier initialisation~\cite{Glorot2010UnderstandingTD}. (4) The activation function $a$ is twice differentiable.

We now derive the training dynamics of the PINNs in two limiting kernels taking account of the feature mapping as the first layer.
\begin{theorem}[Propagation of the Conjugate Kernel]
\label{Prop: ck}
Let input $\mathbf{x}\in \mathbb{R}^{N\times n}$, and each layer of the Neural Network is parameterised with independent and identically distributed (i.i.d.) weights and biases from standard Gaussian distribution. Hence $f^l(\mathbf{x}; \theta_0)\sim \mathcal{GP}(0, \Sigma^{l}(\mathbf{x}, \mathbf{x}'))$, and the Conjugate Kernels propagate through the Neural Network in the following recursive form:
\setlength{\abovedisplayskip}{0pt}
\begin{equation}
\begin{aligned}
 \Sigma^{0}(\mathbf{x}, \mathbf{x}')&= \langle \mathbf{x}, \mathbf{x}'\rangle + 1,\\
 \Sigma^1(\mathbf{x}, \mathbf{x}')&= \mathbf{E}[\varphi(\mathbf{x})^T\varphi(\mathbf{x}')] + 1,\\
 \Sigma^l(\mathbf{x}, \mathbf{x}')&= \mathbf{E}[a(\mathcal{X})^Ta(\mathcal{X}')] + 1, \quad 2 \leq l \leq L,\\
\end{aligned}
\end{equation}
\vskip -0.1in
where $\varphi$ is the feature mapping function at $l=1$ and $\mathcal{X}, \mathcal{X}'$ are the hidden layer state from previous layer and $\left[\begin{array}{cc} \mathcal{X} \\ \mathcal{X}' \end{array}\right]\sim \mathcal{N}  \left( \left[\begin{array}{cc} 0 \\0 \end{array}\right], \left[\begin{array}{cc}\Sigma^{l-1}(\mathbf{x}, \mathbf{x}) & \Sigma^{l-1}\left(\mathbf{x}^{\prime}, \mathbf{x}\right) \\ \Sigma^{l-1}\left(\mathbf{x}, \mathbf{x}^{\prime}\right) & \Sigma^{l-1}\left(\mathbf{x}^{\prime}, \mathbf{x}^{\prime}\right)\end{array}\right] \right)$.\\
\end{theorem}
\vspace{-0.2in}
\begin{proof}
see Appendix \ref{Append: ck}.
\end{proof}
\vspace{-0.1in}
Here we derive the initial distribution of each layer in the network through the feature mapping layer and the non-linear activation layers by computing the explicit Conjugate Kernels. This indicates the Gaussian distribution can propagate layers from the very first feature layer. One important note is that $\mathbf{E}(f^l(\mathbf{x}; \theta_0)) = 0$ holds true for all layers after the feature mapping layer, however, it is only true for the feature mapping layer if the features are randomly sampled. This gives us an insight into the design of feature mapping functions, i.e., $\varphi$ needs to incorporate randomly sampled initialisation. Most importantly, the eigenvalues of the CK embodied the distribution of Collocation points and  IC/BC points and are manipulated by the feature mapping function in the first layer. 

We now investigate the training dynamics of the PINNs by linking the CK and NTK.
\begin{theorem}[Evolution of the NTK with CK]
\label{theorem: ntk}
Let input $\mathbf{x}\in \mathbb{R}^{N\times n}$, $\phi(\mathbf{x})=\varphi(\mathbf{x}) \bigvee a(\mathbf{x})$ ; Recall $\Sigma^1(\mathbf{x}, \mathbf{x}')= \mathbf{E}[\varphi(\mathbf{x})^T\varphi(\mathbf{x}')] + 1$, $\Sigma^l(\mathbf{x},\mathbf{x}') = \mathbf{E}[\phi(\mathbf{x})\phi(\mathbf{x}')]+1$ and its derivative is $\dot{\Sigma}^l(\mathbf{x},\mathbf{x}') = \mathbf{E}[\dot{\phi}(\mathbf{x})\dot{\phi}(\mathbf{x}')], \in  \mathbb{R}^{N \times N}$. Assuming the infinity width limit, the gradient $\nabla f^l$ satisfies:
\setlength{\abovedisplayskip}{0pt}
\begin{equation}
\nabla_\theta f^l(\mathbf{x}; \theta_0)^T\nabla_\theta f^l(\mathbf{x}'; \theta_0) \to \Theta^l(\mathbf{x}, \mathbf{x}'),
\end{equation}
The evolution of the kernels follows:
\begin{equation}
\begin{aligned}
\Theta^1(\mathbf{x}, \mathbf{x}') &= \Theta^{0}(\mathbf{x}, \mathbf{x}')\dot{\Sigma}^1(\mathbf{x}, \mathbf{x}') + \Sigma^1(\mathbf{x}, \mathbf{x}'),\\
\Theta^l(\mathbf{x}, \mathbf{x}') &= \Theta^{l-1}(\mathbf{x}, \mathbf{x}')\dot{\Sigma}^l(\mathbf{x}, \mathbf{x}') + \Sigma^l(\mathbf{x}, \mathbf{x}'), \quad 2 \leq l \leq L,
\end{aligned}
\end{equation}
\end{theorem}
\vspace{-0.2in}
\begin{proof} 
see Appendix \ref{Append: ntk}.
\end{proof}
\vspace{-0.1in}
The second key theoretical result reveals the NTK in each layer depends on the NTK of the last layer and the CK and its derivative $\dot{\Sigma}$ from the current layer. More importantly, the distribution of the feature mapping layer will propagate through the network from layer 2 and onwards. Subsequently, the training dynamics are primarily driven by the leading eigenvalues of the NTK, which ultimately make the overall convergence slow (See Appendix~\ref{Append: Spectral} for more details on how eigenvalues of the NTK affects convergence). This further solidifies the theory backed by~\cite{Wang2020WhenAW}. The CK derivatives of the feature mapping layer play a key role in the evolution. We require the feature mapping to form a kernel which is at least 1st-order differentiable. Furthermore, the range of the spectrum in a kernel decides the generalisability. A narrow spectrum results in limited expressivity, on the other hand, a wide spectrum can produce high frequency aliasing artifacts~\cite{Tancik2020FourierFL}. Hence, a controllable bandwidth is desirable to address different needs.

\begin{figure}[ht]
\begin{center}
\setlength{\belowcaptionskip}{-20pt}
\centerline{\includegraphics[height=2.8cm]{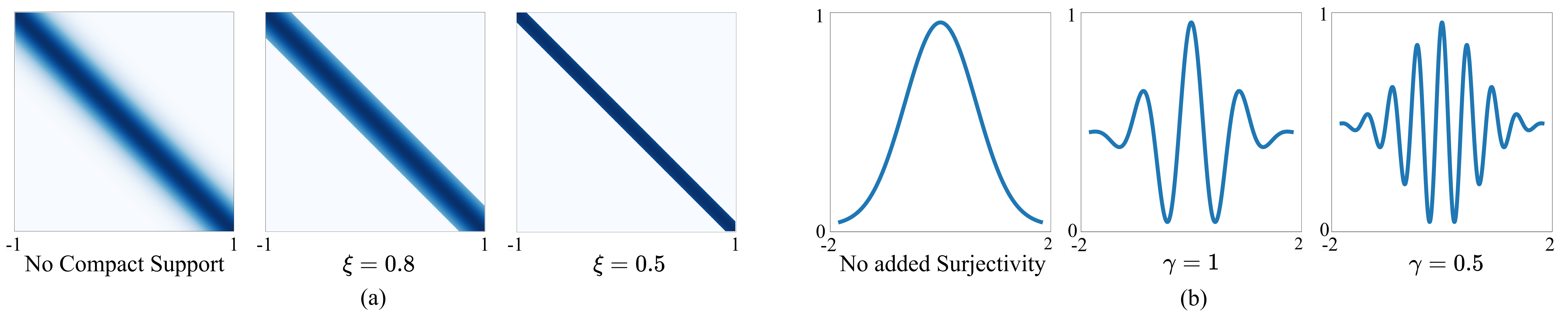}}
\vspace{-0.1in}
\caption{(a) The bandwidth of the kernel can be controlled by the compact support radius of the RBFs; (b) The surjectivity can be adjusted by composing auxiliary Fourier features to the RBFs.}
\label{Fig: rbf support}
\end{center}
\end{figure}

\vspace{-0.1in}
\section{Proposed feature mapping method}
\vspace{-0.1in}
We demonstrated feature mapping is a leading factor in the training dynamics of PINNs. A well-designed feature mapping function can mitigate the spectral bias of the training regime, which should possess the ability to tune the kernel bandwidth and is at least 1st-order differentiable. For high dimensional problems, it is advantageous to compute a less surjective codomain.

Recall that the MLP function is approximated by the convolution of composed NTK $K_{\text{COMP}}=K_{NTK}\circ K_\Phi$ with weighted Dirac delta over the input $x$. We can formulate the $K_{\text{COMP}}$ by:

\vspace{-0.1in}
\begin{equation}
\begin{aligned}
K_{\text{COMP}}(\mathbf{x}) = (K_{\text{COMP}}*\delta_\mathbf{x})(\mathbf{x})
&= \int K_{\text{COMP}}(\mathbf{x}')\delta(\mathbf{x}-\mathbf{x}')dx,\\
&\approx \int K_{\text{COMP}}(\mathbf{x}')K_\Phi(\mathbf{x}-\mathbf{x}')dx,
\end{aligned}
\end{equation}
\vskip -0.1in
The accuracy of the continuous approximation can be analysed by Taylor series expansion:
\begin{equation}
\label{equa: Taylor}
\begin{aligned}
K_{\text{COMP}}(\mathbf{x}) &= \int(K_{\text{COMP}}(\mathbf{x}) + \nabla_{\mathbf{x}}K_{\text{COMP}}(\mathbf{x}-\mathbf{x}') + \frac{1}{2}(\mathbf{x}-\mathbf{x}')\nabla^2 K_{\text{COMP}}(\mathbf{x}-\mathbf{x}')\\
&+ \mathcal{O}((\mathbf{x}-\mathbf{x}')^3))K_\Phi(\mathbf{x}-\mathbf{x}') dx,\\
&= K_{\text{COMP}}(\mathbf{x})\int K_\Phi(\mathbf{x}-\mathbf{x}') dx  +\nabla_{\mathbf{x}}K_{\text{COMP}}(\mathbf{x}-\mathbf{x}')\int(\mathbf{x}-\mathbf{x}') K_\Phi(\mathbf{x}-\mathbf{x}') dx \\
&+\mathcal{O}((\mathbf{x}-\mathbf{x}')^2),
\end{aligned}
\end{equation}
\vskip -0.1in
To make sure the composing kernel is 1st-order accurate, we require the term $\int K_\Phi(\mathbf{x}-\mathbf{x}') d\mathbf{x} = 1$ and the second term in Equation~\ref{equa: Taylor} to be 0. This can be achieved simply by normalising the feature mapping kernel and set a symmetry condition.
We propose a positive definite Radial Basis Function (RBF) for such kernel, and the feature mapping function is given by:
\begin{equation}
\label{equation: rbf}
\Phi(\mathbf{x}) = \frac{w_i\varphi(|\mathbf{x}-c_i|)}{\sum^m_iw_i\varphi(|\mathbf{x}-c_i|)},
\end{equation}
\vskip -0.1in
where $\mathbf{x}\in \mathbb{R}^n$ is the input data, $c \in \mathbb{R}^{m}$ are the centres of the RBFs and are trainable parameters. A natural choice for the RBF is the Gaussian function, $\varphi(\mathbf{x})=e^{-\frac{|\mathbf{x}-\mathbf{c}|^2}{\mathbf{\sigma}^2}}$. If we choose the same number of features as the input size (i.e., $n=m$), this is the same as the RBF interpolation method which gives approximate computation of desired function value by kernel regression. The training input size is often very large in PINNs, it does not scale well in this setting. In our empirical study, we show a few hundred RBFs are sufficient to outperform other types of feature mapping functions. At initialisation, $\mathbf{c}$ is sampled from a standard Gaussian to follow the propagation in Theorem~\ref{Prop: ck}. Additionally, RBFs exhibit injective properties with a normalised input ($\textbf{x}\in[0, 1]$, standard practice in PINNs), as the function on the positive axis decreases monotonically. In principle, we can use many other types of RBF without too many restrictions. some examples are detailed in Table~\ref{tab: RBF types}.

\vspace{-0.1in}
\subsection{Compact support RBF}
\vspace{-0.1in}
\label{sec: RBF compact}
A direct method for tuning the bandwidth is to apply compact support. Here, we introduce compact support RBFs. Traditional RBFs, like the Gaussians, approach zero at infinity but never quite reach it (Global support). By implementing a cut-off distance where distant points yield values of zero, we achieve compact support, Figure~\ref{Fig: rbf support} left. This is formulated by:
\begin{equation}
\label{equation: RBF compact}
\Phi(\mathbf{r}, \xi) = 
\begin{cases}
\varphi(\mathbf{r}, \xi), & \mathbf{r}\le\xi\\
0, & \mathbf{r}>\xi\\
\end{cases},
\end{equation}
Where $\mathbf{r} = |\mathbf{x}-c|$ and $\xi$ is an arbitrary cut-off distance and is proportional to the bandwidth of the kernel.
This ensures that points with a high Euclidean distance do not contribute to the features and makes the resulting feature matrix sparse, which can potentially enhance computationally efficiency. Another way to consider the compact support is as a disconnection between the RBF centres and the computational domain during training. It deactivates some of the RBFs in the neural network, similar to the commonly used Dropout technique~\cite{Srivastava2014DropoutAS}.

\vspace{-0.1in}
\subsection{Conditionally positive definite RBF}
\vspace{-0.1in}
In the infinite-width limit, each layer of the neural network form a linear system. One approach to guarantee a unique solution is by incorporating conditionally positive definite functions through an addition of polynomial terms. The weights function are Lagrange multipliers that enables the constraint of the RBF coefficients. The feature mapping function is modified as:
\begin{equation}
\Phi(\mathbf{x}) = \frac{w^m_i\varphi(|\mathbf{x}-c_i|)}{\sum^m_i w^m_i \varphi(|\mathbf{x}-c_i|)} || w^k_j P(\mathbf{x}),
\end{equation}
Where P is the polynomial function, $||$ is a concatenation.
In the feature mapping layer, the resulting matrix can be represented as:
\begin{equation}
\begin{bmatrix} 
f(\mathbf{x})_1 \\
\vdots \\
f(\mathbf{x})_N \\ 
\end{bmatrix} \rightarrow 
\begin{bmatrix} 
\varphi(\mathbf{r}_1^1) & \hdots & \varphi(\mathbf{r}_1^m) & \mid \mid & 1 & \mathbf{x}_1 & \mathbf{x}^{k} \\
\vdots & \ddots & \vdots& \mid \mid  & \vdots & \vdots & \vdots \\
\varphi(\mathbf{r}_N^1) & \hdots & \varphi(\mathbf{r}_N^m) & \mid \mid & 1 & \mathbf{x}_N & \mathbf{x}_N^{k} \\
\end{bmatrix}
\begin{bmatrix} 
W_m\\
- \\
W_k \\
\end{bmatrix},
\end{equation}
\vskip -0.1in
where $k$ is the order of the polynomials.
Empirically, we find the polynomial term can not only  add greater expressivity to the neural network but can also refine non-linear function approximation, such as the Burgers' Equation and Naiver-Stokes Equation. The extra parameters are low in quantity, it does not add too much computational overhead to the overall network. 
\vspace{-0.1in}
\subsection{Adding Surjectivity to RBF}
\vspace{-0.1in}
We have suggested two important properties of the feature mapping methods, namely the kernel bandwidth controllability and feature space mapping surjectivity.
The two properties guard the expressivity and generalisability of the PINNs, respectively. The bandwidth can be tunned by introducing compact support to the feature mapping function, shown in Equation~\ref{equation: RBF compact}.  We investigate if surjectivity is unfavourable in all occasions. To study the effect of surjectivity of the function, a naive approach is to add Fourier features to the original RBFs, Equation~\ref{equation: rbf}: 
\begin{equation}
\label{equation: surjective rbf}
\Phi(\mathbf{x}) = \frac{w_i\varphi(|\mathbf{x}-c_i|)}{\sum^m_iw_i\varphi(|\mathbf{x}-c_i|)}*\text{cos}(\frac{2\pi \mathbf{x}}{\gamma}),
\end{equation}
\vskip -0.1in
where $\gamma$ is a hyperparameter that regulates the extend of surjectivity added to the feature function, Figure~\ref{Fig: rbf support} (b) shows the behaviour of a Gaussian with added Fourier features.  This results a Gabor-like~\cite{Granlund1978InSO} kernel function. 
In the following section, we conduct experiments comparing our methods to other feature mapping functions on various PDEs, and we carried out an ablation study on tuning the bandwidth and surjectivity.

\vspace{-0.1in}
\section{Empirical Results}
\label{section: experiements}
\vspace{-0.1in}
\subsection{Experimental Setup}
\vspace{-0.1in}
\textbf{Comparison of methods.} We compared our methods with other feature mapping methods for coordinate-based input networks. This includes Fourier-based methods such as Basic Encoding (BE), Positional Encoding (PE), Random Fourier Feature (FF) and Sinusoidal Feature (SF) and Non-Fourier-based ones such as Complex Triangle (CT) and Complex Gaussian (CG). The exact function and related literature can be found in Appendix~\ref{Append: FM}. RBF-INT is \textbf{our} standard feature mapping function in the formulation of RBF interpolants. RBF-POL and RBF-COM stand for RBF-INT with polynomials and RBF-INT with Compact Support respectively throughout the paper. We use Gaussian RBF for the main experiments unless otherwise stated.

\textbf{Benchmarked PDEs.} We conducted benchmarks from existing literature~\cite{lu2021deepxde, Hao2023PINNacleAC} on various PDEs in both forward and inverse problems. The forward problems demonstrated include the Wave equation (hyperbolic), Diffusion\&Heat equation (parabolic), Poisson equation (elliptic) and Burgers'\&Navier-Stokes (NS) equations (non-linear). The inverse problems are the Inverse Burgers' equation and Inverse Lorenz equations.  The equations and their boundary conditions are specified in Appendix \ref{Append: Benchmark}.

Implementation details and evaluation method are included in Appendix~\ref{appendix: reproducibility}.

\vspace{-0.1in}
\subsection{Forward Problems}
\vspace{-0.1in}

\begin{table}[h]
\caption{PDEs benchmark results comparing feature mapping methods in $\ell^2$ error.  The best results are in \colorbox{blue!25}{Blue}. Complete experimental results with standard deviations are shown in Appendix~\ref{Append: full results}.}
\label{tab: forward pdes}
\begin{adjustbox}{width=\columnwidth,center}
\begin{tabular}{@{}cccccccccc@{}}
\toprule
& PINN                           & BE                 & PE            & FF                & SF             & CT                     & CG                       & \textbf{RBF-INT}                            & \textbf{RBF-POL}   \\ \midrule
\multicolumn{1}{l|}{Wave}      & \multicolumn{1}{l|}{3.731e-1} & \multicolumn{1}{l|}{1.035e0}  & \multicolumn{1}{l|}{1.014e0}  & \multicolumn{1}{l|}{\colorbox{blue!25}{2.375e-3}} & \multicolumn{1}{l|}{7.932e-3} & \multicolumn{1}{l|}{1.114e0}  & \multicolumn{1}{l|}{1.036e0}  & \multicolumn{1}{l|}{2.814e-2} & 2.361e-2  \\
\multicolumn{1}{l|}{Diffusion} & \multicolumn{1}{l|}{1.426e-4} & \multicolumn{1}{l|}{1.575e-1} & \multicolumn{1}{l|}{1.595e-1} & \multicolumn{1}{l|}{2.334e-3} & \multicolumn{1}{l|}{3.474e-4} & \multicolumn{1}{l|}{1.860e0}  & \multicolumn{1}{l|}{2.721e-2} & \multicolumn{1}{l|}{3.066e-4} & \colorbox{blue!25}{3.498e-5} \\
\multicolumn{1}{l|}{Heat}      & \multicolumn{1}{l|}{4.732e-3} & \multicolumn{1}{l|}{6.491e-3} & \multicolumn{1}{l|}{7.574e-3} & \multicolumn{1}{l|}{2.190e-3} & \multicolumn{1}{l|}{3.961e-3} & \multicolumn{1}{l|}{4.524e-1} & \multicolumn{1}{l|}{2.626e-1}  & \multicolumn{1}{l|}{1.157e-3} & \colorbox{blue!25}{4.098e-4} \\
\multicolumn{1}{l|}{Poisson}   & \multicolumn{1}{l|}{3.618e-3} & \multicolumn{1}{l|}{4.964e-1} & \multicolumn{1}{l|}{4.910e-1} & \multicolumn{1}{l|}{7.586e-4} & \multicolumn{1}{l|}{9.078e-4} & \multicolumn{1}{l|}{6.348e-1} & \multicolumn{1}{l|}{2.334e-1} & \multicolumn{1}{l|}{\colorbox{blue!25}{5.259e-4}} & 8.942e-4 \\
\multicolumn{1}{l|}{Burgers'}   & \multicolumn{1}{l|}{1.864e-3} & \multicolumn{1}{l|}{5.585e-1} & \multicolumn{1}{l|}{5.363e-1} & \multicolumn{1}{l|}{7.496e-2} & \multicolumn{1}{l|}{1.299e-3} & \multicolumn{1}{l|}{9.935e-1} & \multicolumn{1}{l|}{7.521e-1} & \multicolumn{1}{l|}{2.945e-3} & \colorbox{blue!25}{3.159e-4} \\
\multicolumn{1}{l|}{Steady NS} & \multicolumn{1}{l|}{5.264e-1} & \multicolumn{1}{l|}{7.143e-1} & \multicolumn{1}{l|}{6.332e-1} & \multicolumn{1}{l|}{6.939e-1} & \multicolumn{1}{l|}{3.769e-1} & \multicolumn{1}{l|}{5.460e-1}  & \multicolumn{1}{l|}{4.867e-1} & \multicolumn{1}{l|}{2.991e-1} & \colorbox{blue!25}{2.567e-1} \\
\midrule
\end{tabular}
\end{adjustbox}
\vspace{-0.15in}
\end{table}

\textbf{Time-dependent PDEs.} Some benchmarked time-dependent PDEs only have Dirichlet initial conditions (e.g. the Diffusion equation and the Heat equation in Table~\ref{tab: forward pdes}), then their initial condition can be treated as a special type of boundary condition during loss optimisation. This offers us the advantage of homogeneously sampling IC/BC points across spatial and temporal domains. A higher penalty on the IC/BC terms is adopted in the experiments, that is setting $\lambda_r=1$, $\lambda_{ic}=100$ and $\lambda_{bc}=100$ from Equation~\ref{equation: loss}. By doing so, we found it is easier for the solutions from the IC to propagate to the domain, and comply with the hard BC constraints.

Our solution in the Diffusion equation shows superior performance over other methods by some order of magnitude. The boundary errors are visibly higher in Fourier-based methods shown in Figure~\ref{fig:diffusion qualitative}. 

The RBFs are better at handling multiscale problems demonstrated by the Heat equation (~\ref{equation: heat}). The tested Heat equation is stronly directionally anisotropic with coefficients, $\frac{1}{{500\pi}^2}$ and $\frac{1}{{\pi}^2}$ in the x and y directions. Figure~\ref{fig:heat qualitative} has shown our methods preserves the details of the solution in each time step.

\textbf{Non-linear PDEs.} We evaluate the methods on two classic non-linear PDEs, the Burgers' equation and the Navier-Stokes equation. It has been shown in Figure~\ref{fig:burgers qualitative} that the RBFs with polynomial terms are more capable of solving the discontinuity at $x=0$. The steady N-S equation has no time derivative term. However, the back step flow geometry makes the model harder to generalise, hence we again penalise the BC loss term with a magnitude of 100. Our methods achieve higher accuracy.

All cases are tested with 2k sampling points at each boundary and 20k collocation points in the domain. The Wave equation has an addition Neumann boundary condition that is treated by the differential operation like the PDEs but added to the IC loss term.

\vspace{-0.1in}
\subsection{Inverse Problems}
\vspace{-0.1in}
A major application of the PINNs is their ability to solve inverse problems. The unknown coefficients in the differential equations can be discovered by a small amount of data points, take the Lorenz system (\ref{equation: lorenz}) as an example, 
the $\alpha$, $\rho$ and $\beta$ are three unknown coefficients during training. We can explicitly attach the coefficients to the neural network as learnable external parameters. The coefficients along with the PDEs are to construct the PDEs loss. Thereafter, the model is able to converge and the learnable external parameters are optimised to determine the ideal coefficients. 

In the inverse Burgers' equation problem, there are 5000 data points used and the same number of randomly sampled points are used to compute the PDE loss. For the inverse Lorenz system, only 40 data points are used and 400 collocations are sampled in $t \in [0, 3]$. Lorenz system is sensitive to coefficients and initial position changes. The initial positions $x_0 = 0$, $y_0 = 1$, $z_0 = 1.05$ are not provided to the model.

 Another experiment conducted is to test if the feature mapping functions are prone to noise. $1\%$ Gaussian noises are added to the inverse Burgers' problem and $0.5\%$ to the Lorenz system data. The results shown in Table~\ref{tab: inverse} indicate the 4 feature mapping methods tested are robust to noises to some degree. Overall, RBF-POL is the most resilient feature mapping function to noises. 
 
 \vskip -0.2in
 \begin{table}[ht]
\caption{Benchmark on the inverse problems in $\ell^2$ error. * indicates problems with noises added to the data. Full results with mean and standard deviations in Appendix~\ref{Append: full results}.}
\label{tab: inverse}
\footnotesize
\centering
\begin{tabular}{ccccc}
\toprule 
& FF                             & SF                            & \textbf{RBF-INT}                       & \textbf{RBF-POL}  \\ \midrule
\multicolumn{1}{l|}{I-Burgers'}  & \multicolumn{1}{l|}{2.391e-2} & \multicolumn{1}{l|}{2.436e-2} & \multicolumn{1}{l|}{1.741e-2} & \colorbox{blue!25}{1.575e-2} \\
\multicolumn{1}{l|}{I-Lorenz}   & \multicolumn{1}{l|}{6.516e-3}  & \multicolumn{1}{l|}{6.390e-3} & \multicolumn{1}{l|}{6.080e-3} & \colorbox{blue!25}{5.991e-3} \\ 
\multicolumn{1}{l|}{I-Burgers'*}  & \multicolumn{1}{l|}{2.509e-2} & \multicolumn{1}{l|}{2.913e-2} & \multicolumn{1}{l|}{1.993e-2} & \colorbox{blue!25}{1.753e-2} \\
\multicolumn{1}{l|}{I-Lorenz*}   & \multicolumn{1}{l|}{7.934e-3}  & \multicolumn{1}{l|}{6.856e-3} & \multicolumn{1}{l|}{6.699e-3} & \colorbox{blue!25}{6.342e-3} \\ 
\midrule
\end{tabular}
\end{table}

\begin{figure}[ht]
\begin{center}
\setlength{\belowcaptionskip}{-30pt}
\centerline{\includegraphics[height=3.5cm]{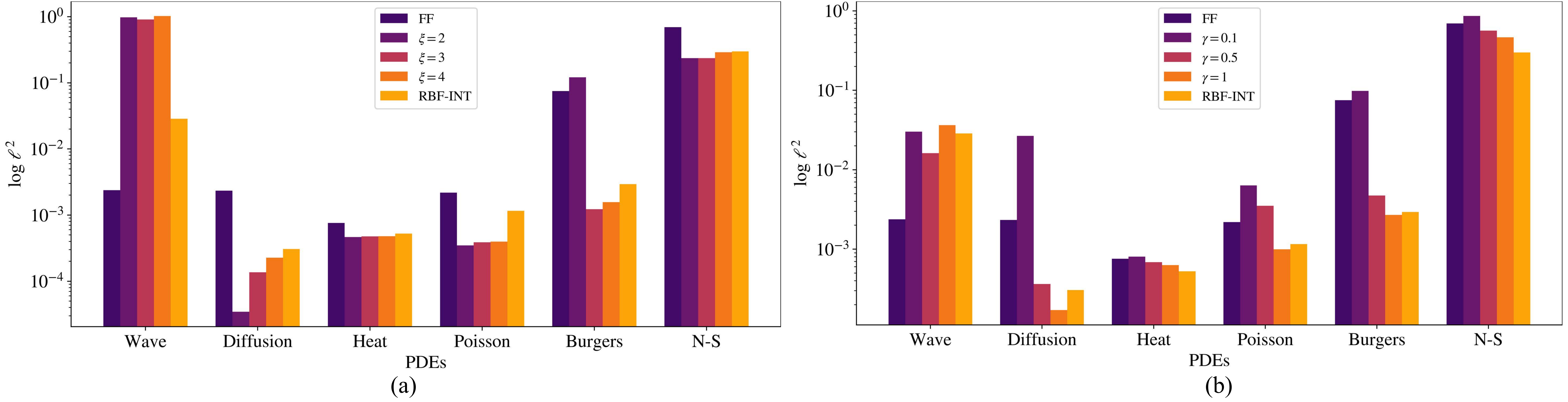}}
\vspace{-0.1in}
\caption{(a) Although the optimal value of $\xi$ varies, narrower kernel bandwidths are generally preferred in some PDEs. (b) Adding too much surjectivity to RBF is mostly disadvantageous, but it can bring extra performance improvements in a few situations.}
\label{figure: support and surjectivity}
\end{center}
\end{figure}


\vspace{-0.2in}
\subsection{Ablation study}
\vspace{-0.1in}
Here, we present experimental results using compact support RBF with different cut-off distance $\xi$ and RBF with added surjectivity controlled by $\gamma$. In Figure~\ref{figure: support and surjectivity} (a),  it is noticeable that the performance of RBF generally improves with shorter support, while the Wave Equation does not converge with narrow kernels.  We found that the optimal value of $\xi$ differ for best performance in each PDE, which may be due to different domain limits and boundary conditions. Overall, it is adverse to have too much added artificial surjectivity on the top of RBFs. For some Equations, the composed feature mapping function tends to generalised better than the plain ones, we attribute this to the better expressivity-generalisability trade-off. We also observe that the two hyperparameters are not contradictory. In the Diffusion equation, both narrow compact support and additional surjectivity improve overall performance. The nuance of balancing the two comes down to individual implications.

\vspace{-0.05in}
\textbf{Other experimental results (Appendix~\ref{appendix: ablation study}).} We study the performance of RBFs with different settings. The number of RBFs generally has a positive impact on reducing error (Figure~\ref{ablation: rbf number}). This is also true for the number of polynomial terms (Figure~\ref{Fig: abla-pol}). We found that 128 RBF features with 10 polynomial terms yield good results without excessive computational overhead. Among the five types of RBFs we tested, Gaussian RBFs achieved the lowest error in all PDEs (Figure~\ref{Fig: RBF types}). 

\vspace{-0.05in}
\textbf{Convergence, Complexity and Scalability (Appendix~\ref{appendix: complexity}).} Our feature mapping method helps PINNs to converge faster in some Equations, as demonstrated in Figure~\ref{figure: convergence}. With additional Polynomial terms, there is an auxiliary 10k parameters on the top of the RBF features. This is mostly negligible in modern GPU speed. We include Table~\ref{tab: complexity} to show the complexity of feature mapping layer with a standard MLP.
A test on different amount of sample points are demonstrated in~\ref{Fig: abla-pol}, we conclude all feature mapping methods can scale relatively well, but software optimisation can vary case by case.

\vspace{-0.15in}
\section{Limitations and Future work}
\label{section: limitation}
\vspace{-0.15in}
Our theoretical work was carried out primarily on MLP based PINNs, and its adoption in Physics Informed Convolution Neural Network~\cite{Wandel2021SplinePINNAP} and Physics Informed Transformers~\cite{Zhao2023PINNsFormerAT} is readily attainable. The theoretical hypothesis is under the Neural Tangent Kernel limit and a few assumptions have been made. In the proof, the other choice of neural network components such as the periodic activation functions are not yet investigated. This leaves us an exciting research path to future work. Our proposed feature mapping method inevitably suffers from the curse of dimensionality like other exiting feature mapping methods. That is when the dimensions of the PDEs are very high, feature embedders will require a corresponding large scale. On the other hand, there will be more RBF functions needed in extremely high frequency PDEs, such that Fourier features would possibly be a good alternative. Our theory did not provide thorough guidelines on balancing the hyperparameters $\xi$ and $\gamma$ for all physics problems. Composing other feature mapping methods, i,e., fine tuning the bandwidth and surjectivity, can be a compelling approach to solve more complex problems.
\vspace{-0.15in}
\section{Conclusion}
\vspace{-0.15in}
In conclusion, we provided theoretical proof that feature mapping in PINNs influences the Conjugate Kernel and Neural Tanget Kernel which dominate the training dynamic of PINNs.  We introduce a framework to design an effective feature mapping function in PINNs and propose Radial Basis Function based feature mapping approaches. Our method not only improves the generalisation in a range of forward and inverse physics problems but also outperforms other feature mapping methods by a significant margin. RBF feature mapping can potentially work with many other PINNs techniques such as some novel activation functions and different types of loss or training strategies such as curriculum training. While this work focuses on solving PDEs, RBF feature mapping continues to explore its application in other coordinates-based input neural networks for different tasks.

\newpage
\medskip

{
\small
\bibliography{main_cite}

\begin{thebibliography}{64}
\providecommand{\natexlab}[1]{#1}
\providecommand{\url}[1]{\texttt{#1}}
\expandafter\ifx\csname urlstyle\endcsname\relax
  \providecommand{\doi}[1]{doi: #1}\else
  \providecommand{\doi}{doi: \begingroup \urlstyle{rm}\Url}\fi

\bibitem[Karniadakis et~al.(2021)Karniadakis, Kevrekidis, Lu, Perdikaris, Wang, and Yang]{Karniadakis2021PhysicsinformedML}
George~Em Karniadakis, Ioannis~G. Kevrekidis, Lu~Lu, Paris Perdikaris, Sifan Wang, and Liu Yang.
\newblock Physics-informed machine learning.
\newblock \emph{Nature Reviews Physics}, 3:\penalty0 422 -- 440, 2021.

\bibitem[Smith et~al.(2022)Smith, Seccamonte, Swami, and Bullo]{Smith2022PhysicsInformedIR}
Kevin~D. Smith, Francesco Seccamonte, Ananthram Swami, and Francesco Bullo.
\newblock Physics-informed implicit representations of equilibrium network flows.
\newblock In \emph{Neural Information Processing Systems}, 2022.

\bibitem[Hu et~al.(2023)Hu, Li, Klemme, Nam, Ma, Amrouch, and Xiong]{Hu2023SyncTREE}
Yuting Hu, Jiajie Li, Florian Klemme, Gi-Joon Nam, Tengfei Ma, Hussam Amrouch, and Jinjun Xiong.
\newblock Synctree: Fast timing analysis for integrated circuit design through a physics-informed tree-based graph neural network.
\newblock In \emph{Neural Information Processing Systems}, 2023.

\bibitem[Nicoli et~al.(2023)Nicoli, Anders, Funcke, Hartung, Jansen, Kuhn, Muller, Stornati, Kessel, and Nakajima]{Nicoli2023Quant}
Kim~Andrea Nicoli, Christopher~J Anders, Lena Funcke, Tobias Hartung, Karl Jansen, Stefan Kuhn, Klaus~Robert Muller, Paolo Stornati, Pan Kessel, and Shinichi Nakajima.
\newblock Physics-informed bayesian optimization of variational quantum circuits.
\newblock In \emph{Neural Information Processing Systems}, 2023.

\bibitem[Goyeneche et~al.(2023)Goyeneche, Ramachandran, Wang, Karasan, Cheng, Stella, and Lustig]{Goyeneche2023ResoNet}
Alfredo~De Goyeneche, Shreya Ramachandran, Ke~Wang, Ekin Karasan, Joseph~Yitan Cheng, X~Yu Stella, and Michael Lustig.
\newblock Resonet: Noise-trained physics-informed mri off-resonance correction.
\newblock In \emph{Neural Information Processing Systems}, 2023.

\bibitem[Salehi and Giannacopoulos(2022)]{Salehi2021PhysGNNAP}
Yasmin Salehi and Dennis~D. Giannacopoulos.
\newblock Physgnn: A physics-driven graph neural network based model for predicting soft tissue deformation in image-guided neurosurgery.
\newblock In \emph{Neural Information Processing Systems}, 2022.

\bibitem[Pokkunuru et~al.(2023)Pokkunuru, Rooshenas, Strauss, Abhishek, and Khan]{pokkunuru2023improved}
Akarsh Pokkunuru, Amirmohmmad Rooshenas, Thilo Strauss, Anuj Abhishek, and Taufiquar Khan.
\newblock Improved training of physics-informed neural networks using energy-based priors: a study on electrical impedance tomography.
\newblock In \emph{International Conference on Learning Representations}, 2023.

\bibitem[Thangamuthu et~al.(2022)Thangamuthu, Kumar, Bishnoi, Bhattoo, Krishnan, and Ranu]{Thangamuthu2022Dynamical}
Abishek Thangamuthu, Gunjan Kumar, Suresh Bishnoi, Ravinder Bhattoo, N~M~Anoop Krishnan, and Sayan Ranu.
\newblock Unravelling the performance of physics-informed graph neural networks for dynamical systems.
\newblock In \emph{Neural Information Processing Systems}, 2022.

\bibitem[Ni and Qureshi(2023)]{Ni2022NTFieldsNT}
Ruiqi Ni and Ahmed~Hussain Qureshi.
\newblock Ntfields: Neural time fields for physics-informed robot motion planning.
\newblock In \emph{International Conference on Learning Representations}, 2023.

\bibitem[Kashinath et~al.(2021)Kashinath, Mustafa, Albert, Wu, Jiang, Esmaeilzadeh, Azizzadenesheli, Wang, Chattopadhyay, Singh, et~al.]{kashinath2021physics}
Karthik Kashinath, M~Mustafa, Adrian Albert, JL~Wu, C~Jiang, Soheil Esmaeilzadeh, Kamyar Azizzadenesheli, R~Wang, A~Chattopadhyay, A~Singh, et~al.
\newblock Physics-informed machine learning: case studies for weather and climate modelling.
\newblock \emph{Philosophical Transactions of the Royal Society A}, 379\penalty0 (2194):\penalty0 20200093, 2021.

\bibitem[Giladi et~al.(2021)Giladi, Ben-Haim, Nevo, Matias, and Soudry]{Giladi2021PhysicsAwareDW}
Niv Giladi, Zvika Ben-Haim, Sella Nevo, Yossi Matias, and Daniel Soudry.
\newblock Physics-aware downsampling with deep learning for scalable flood modeling.
\newblock In \emph{Neural Information Processing Systems}, 2021.

\bibitem[Raissi et~al.(2019)Raissi, Perdikaris, and Karniadakis]{Raissi2019PhysicsinformedNN}
Maziar Raissi, Paris Perdikaris, and George~Em Karniadakis.
\newblock Physics-informed neural networks: A deep learning framework for solving forward and inverse problems involving nonlinear partial differential equations.
\newblock \emph{J. Comput. Phys.}, 378:\penalty0 686--707, 2019.

\bibitem[Xie et~al.(2021)Xie, Takikawa, Saito, Litany, Yan, Khan, Tombari, Tompkin, Sitzmann, and Sridhar]{Xie2021NeuralFI}
Yiheng Xie, Towaki Takikawa, Shunsuke Saito, Or~Litany, Shiqin Yan, Numair Khan, Federico Tombari, James Tompkin, Vincent Sitzmann, and Srinath Sridhar.
\newblock Neural fields in visual computing and beyond.
\newblock \emph{Computer Graphics Forum}, 41, 2021.

\bibitem[Wang et~al.(2020)Wang, Yu, and Perdikaris]{Wang2020WhenAW}
Sifan Wang, Xinling Yu, and Paris Perdikaris.
\newblock When and why pinns fail to train: A neural tangent kernel perspective.
\newblock \emph{J. Comput. Phys.}, 449:\penalty0 110768, 2020.

\bibitem[Geiger et~al.(2019)Geiger, Spigler, Jacot, and Wyart]{Geiger2019DisentanglingFA}
Mario Geiger, Stefano Spigler, Arthur Jacot, and Matthieu Wyart.
\newblock Disentangling feature and lazy training in deep neural networks.
\newblock \emph{Journal of Statistical Mechanics: Theory and Experiment}, 2019.

\bibitem[Krishnapriyan et~al.(2021)Krishnapriyan, Gholami, Zhe, Kirby, and Mahoney]{Krishnapriyan2021CharacterizingPF}
Aditi~S. Krishnapriyan, Amir Gholami, Shandian Zhe, Robert~M. Kirby, and Michael~W. Mahoney.
\newblock Characterizing possible failure modes in physics-informed neural networks.
\newblock In \emph{Neural Information Processing Systems}, 2021.

\bibitem[Daw et~al.(2022)Daw, Bu, Wang, Perdikaris, and Karpatne]{Daw22Propagation}
Arka Daw, Jie Bu, Sifan Wang, Paris Perdikaris, and Anuj Karpatne.
\newblock Mitigating propagation failures in physics-informed neural networks using retain-resample-release (r3) sampling.
\newblock In \emph{International Conference on Machine Learning}, 2022.

\bibitem[Wang et~al.(2021{\natexlab{a}})Wang, Teng, and Perdikaris]{Wang2021UnderstandingAM}
Sifan Wang, Yujun Teng, and Paris Perdikaris.
\newblock Understanding and mitigating gradient flow pathologies in physics-informed neural networks.
\newblock \emph{SIAM J. Sci. Comput.}, 43:\penalty0 A3055--A3081, 2021{\natexlab{a}}.

\bibitem[Wang et~al.(2021{\natexlab{b}})Wang, Wang, and Perdikaris]{WANG2021113938}
Sifan Wang, Hanwen Wang, and Paris Perdikaris.
\newblock On the eigenvector bias of fourier feature networks: From regression to solving multi-scale pdes with physics-informed neural networks.
\newblock \emph{Computer Methods in Applied Mechanics and Engineering}, 2021{\natexlab{b}}.

\bibitem[Wang et~al.(2022{\natexlab{a}})Wang, Sankaran, and Perdikaris]{Wang2022RespectingCI}
Sifan Wang, Shyam Sankaran, and Paris Perdikaris.
\newblock Respecting causality is all you need for training physics-informed neural networks.
\newblock \emph{ArXiv}, abs/2203.07404, 2022{\natexlab{a}}.

\bibitem[Psaros et~al.(2021)Psaros, Kawaguchi, and Karniadakis]{Psaros2021MetalearningPL}
Apostolos~F. Psaros, Kenji Kawaguchi, and George~Em Karniadakis.
\newblock Meta-learning pinn loss functions.
\newblock \emph{J. Comput. Phys.}, 458:\penalty0 111121, 2021.

\bibitem[Jagtap and Karniadakis(2020)]{Jagtap2020ExtendedPN}
Ameya~Dilip Jagtap and George~E. Karniadakis.
\newblock Extended physics-informed neural networks (xpinns): A generalized space-time domain decomposition based deep learning framework for nonlinear partial differential equations.
\newblock \emph{Communications in Computational Physics}, 2020.

\bibitem[Kharazmi et~al.(2020)Kharazmi, Zhang, and Karniadakis]{Kharazmi2020hpVPINNsVP}
Ehsan Kharazmi, Zhongqiang Zhang, and George~Em Karniadakis.
\newblock hp-vpinns: Variational physics-informed neural networks with domain decomposition.
\newblock \emph{ArXiv}, abs/2003.05385, 2020.

\bibitem[Moseley et~al.(2021)Moseley, Markham, and Nissen‐Meyer]{Moseley2021FiniteBP}
Benjamin Moseley, A.~Markham, and Tarje Nissen‐Meyer.
\newblock Finite basis physics-informed neural networks (fbpinns): a scalable domain decomposition approach for solving differential equations.
\newblock \emph{Advances in Computational Mathematics}, 49, 2021.

\bibitem[Li et~al.(2022)Li, Penwarden, Kirby, and Zhe]{Li2022MetaLO}
Shibo Li, Michael Penwarden, Robert~M. Kirby, and Shandian Zhe.
\newblock Meta learning of interface conditions for multi-domain physics-informed neural networks.
\newblock In \emph{International Conference on Machine Learning}, 2022.

\bibitem[Yu et~al.(2022)Yu, Lu, Meng, and Karniadakis]{Yu2022Gradient}
Jeremy Yu, Lu~Lu, Xuhui Meng, and George Karniadakis.
\newblock Gradient-enhanced physics-informed neural networks for forward and inverse pde problems.
\newblock \emph{Computer Methods in Applied Mechanics and Engineering}, 393:\penalty0 114823, 03 2022.

\bibitem[Wang et~al.(2022{\natexlab{b}})Wang, Li, He, and Wang]{Wang22l2}
Chuwei Wang, Shanda Li, Di~He, and Liwei Wang.
\newblock Is l2 physics informed loss always suitable for training physics informed neural network?
\newblock In \emph{Neural Information Processing Systems}, 2022{\natexlab{b}}.

\bibitem[Akhound-Sadegh et~al.(2023)Akhound-Sadegh, Perreault-Levasseur, Brandstetter, Welling, and Ravanbakhsh]{AkhoundSadegh2023LiePS}
Tara Akhound-Sadegh, Laurence Perreault-Levasseur, Johannes Brandstetter, Max Welling, and Siamak Ravanbakhsh.
\newblock Lie point symmetry and physics informed networks.
\newblock In \emph{Neural Information Processing Systems}, 2023.

\bibitem[Wong et~al.(2022)Wong, Ooi, Gupta, and Ong]{Wong2022Sinusoidal}
Jian Wong, Chinchun Ooi, Abhishek Gupta, and Yew Ong.
\newblock Learning in sinusoidal spaces with physics-informed neural networks.
\newblock \emph{IEEE Transactions on Artificial Intelligence}, PP:\penalty0 1--5, 01 2022.
\newblock \doi{10.1109/TAI.2022.3192362}.

\bibitem[Tancik et~al.(2020)Tancik, Srinivasan, Mildenhall, Fridovich-Keil, Raghavan, Singhal, Ramamoorthi, Barron, and Ng]{Tancik2020FourierFL}
Matthew Tancik, Pratul~P. Srinivasan, Ben Mildenhall, Sara Fridovich-Keil, Nithin Raghavan, Utkarsh Singhal, Ravi Ramamoorthi, Jonathan~T. Barron, and Ren Ng.
\newblock Fourier features let networks learn high frequency functions in low dimensional domains.
\newblock In \emph{Neural Information Processing Systems}, 2020.

\bibitem[Hu and Huang(2021)]{Hu2021OnTR}
Zhengmian Hu and Heng Huang.
\newblock On the random conjugate kernel and neural tangent kernel.
\newblock In \emph{International Conference on Machine Learning}, 2021.

\bibitem[Gao et~al.(2023)Gao, Gu, and Ng]{Gao2023GradientDF}
Yihang Gao, Yiqi Gu, and Michael~K. Ng.
\newblock Gradient descent finds the global optima of two-layer physics-informed neural networks.
\newblock In \emph{International Conference on Machine Learning}, 2023.

\bibitem[Fan and Wang(2020)]{Fan2020SpectraOT}
Zhou Fan and Zhichao Wang.
\newblock Spectra of the conjugate kernel and neural tangent kernel for linear-width neural networks.
\newblock In \emph{Neural Information Processing Systems}, 2020.

\bibitem[Jacot et~al.(2018)Jacot, Gabriel, and Hongler]{Jacot2018NeuralTK}
Arthur Jacot, Franck Gabriel, and Cl{\'e}ment Hongler.
\newblock Neural tangent kernel: Convergence and generalization in neural networks.
\newblock In \emph{Neural Information Processing Systems}, 2018.

\bibitem[Sitzmann et~al.(2020)Sitzmann, Martel, Bergman, Lindell, and Wetzstein]{Sitzmann2020ImplicitNR}
Vincent Sitzmann, Julien N.~P. Martel, Alexander~W. Bergman, David~B. Lindell, and Gordon Wetzstein.
\newblock Implicit neural representations with periodic activation functions.
\newblock In \emph{Neural Information Processing Systems}, 2020.

\bibitem[Ramasinghe and Lucey(2021)]{Ramasinghe2021ALR}
Sameera Ramasinghe and Simon Lucey.
\newblock A learnable radial basis positional embedding for coordinate-mlps.
\newblock In \emph{AAAI Conference on Artificial Intelligence}, 2021.

\bibitem[Lee et~al.(2019)Lee, Xiao, Schoenholz, Bahri, Novak, Sohl-Dickstein, and Pennington]{Lee2019WideNN}
Jaehoon Lee, Lechao Xiao, Samuel~S. Schoenholz, Yasaman Bahri, Roman Novak, Jascha~Narain Sohl-Dickstein, and Jeffrey Pennington.
\newblock Wide neural networks of any depth evolve as linear models under gradient descent.
\newblock In \emph{Neural Information Processing Systems}, 2019.

\bibitem[Glorot and Bengio(2010)]{Glorot2010UnderstandingTD}
Xavier Glorot and Yoshua Bengio.
\newblock Understanding the difficulty of training deep feedforward neural networks.
\newblock In \emph{International Conference on Artificial Intelligence and Statistics}, 2010.

\bibitem[Srivastava et~al.(2014)Srivastava, Hinton, Krizhevsky, Sutskever, and Salakhutdinov]{Srivastava2014DropoutAS}
Nitish Srivastava, Geoffrey~E. Hinton, Alex Krizhevsky, Ilya Sutskever, and Ruslan Salakhutdinov.
\newblock Dropout: a simple way to prevent neural networks from overfitting.
\newblock \emph{J. Mach. Learn. Res.}, 15:\penalty0 1929--1958, 2014.

\bibitem[Granlund(1978)]{Granlund1978InSO}
Gösta~H Granlund.
\newblock In search of a general picture processing operator.
\newblock \emph{Computer Graphics and Image Processing}, 8:\penalty0 155--173, 1978.

\bibitem[Lu et~al.(2021)Lu, Meng, Mao, and Karniadakis]{lu2021deepxde}
Lu~Lu, Xuhui Meng, Zhiping Mao, and George~Em Karniadakis.
\newblock {DeepXDE}: A deep learning library for solving differential equations.
\newblock \emph{SIAM Review}, 63\penalty0 (1):\penalty0 208--228, 2021.
\newblock \doi{10.1137/19M1274067}.

\bibitem[Hao et~al.(2023)Hao, Yao, Su, Su, Wang, Lu, Xia, Zhang, Liu, Lu, and Zhu]{Hao2023PINNacleAC}
Zhongkai Hao, Jiachen Yao, Chang Su, Hang Su, Ziao Wang, Fanzhi Lu, Zeyu Xia, Yichi Zhang, Songming Liu, Lu~Lu, and Jun Zhu.
\newblock Pinnacle: A comprehensive benchmark of physics-informed neural networks for solving pdes.
\newblock \emph{ArXiv}, abs/2306.08827, 2023.

\bibitem[Wandel et~al.(2021)Wandel, Weinmann, Neidlin, and Klein]{Wandel2021SplinePINNAP}
Nils Wandel, Michael Weinmann, Michael Neidlin, and R.~Klein.
\newblock Spline-pinn: Approaching pdes without data using fast, physics-informed hermite-spline cnns.
\newblock \emph{ArXiv}, abs/2109.07143, 2021.

\bibitem[Zhao et~al.(2024)Zhao, Ding, and Prakash]{Zhao2023PINNsFormerAT}
Leo Zhao, Xueying Ding, and B.~Aditya Prakash.
\newblock Pinnsformer: A transformer-based framework for physics-informed neural networks.
\newblock In \emph{International Conference on Learning Representations}, 2024.

\bibitem[Das and Tesfamariam(2022)]{Das2022StateoftheArtRO}
Sourav Das and Solomon Tesfamariam.
\newblock State-of-the-art review of design of experiments for physics-informed deep learning.
\newblock \emph{ArXiv}, abs/2202.06416, 2022.

\bibitem[Wu et~al.(2023)Wu, Zhu, Tan, Kartha, and Lu]{WU2023115671}
Chenxi Wu, Min Zhu, Qinyang Tan, Yadhu Kartha, and Lu~Lu.
\newblock A comprehensive study of non-adaptive and residual-based adaptive sampling for physics-informed neural networks.
\newblock \emph{Computer Methods in Applied Mechanics and Engineering}, 403:\penalty0 115671, 2023.
\newblock ISSN 0045-7825.

\bibitem[Nabian et~al.(2021)Nabian, Gladstone, and Meidani]{Nabian2021EfficientTO}
Mohammad~Amin Nabian, Rini~Jasmine Gladstone, and Hadi Meidani.
\newblock Efficient training of physics‐informed neural networks via importance sampling.
\newblock \emph{Computer‐Aided Civil and Infrastructure Engineering}, 36:\penalty0 962 -- 977, 2021.

\bibitem[Yang et~al.(2023)Yang, Qiu, and Fu]{Yang23DMIS}
Zijiang Yang, Zhongwei Qiu, and Dongmei Fu.
\newblock Dmis: Dynamic mesh-based importance sampling for training physics-informed neural networks.
\newblock AAAI Press, 2023.

\bibitem[Lau et~al.(2024)Lau, Hemachandra, Ng, and Low]{Lau2024PINNACLEPA}
Gregory Kang~Ruey Lau, Apivich Hemachandra, See-Kiong Ng, and Bryan Kian~Hsiang Low.
\newblock Pinnacle: Pinn adaptive collocation and experimental points selection.
\newblock In \emph{International Conference on Learning Representations}, 2024.

\bibitem[Jagtap and Karniadakis(2019)]{Jagtap2019AdaptiveAF}
Ameya~Dilip Jagtap and George~Em Karniadakis.
\newblock Adaptive activation functions accelerate convergence in deep and physics-informed neural networks.
\newblock \emph{J. Comput. Phys.}, 404, 2019.

\bibitem[Jagtap et~al.(2020)Jagtap, Kawaguchi, and Karniadakis]{Jagtap2020LocallyAA}
Ameya~Dilip Jagtap, Kenji Kawaguchi, and George~Em Karniadakis.
\newblock Locally adaptive activation functions with slope recovery for deep and physics-informed neural networks.
\newblock \emph{Proceedings of the Royal Society A}, 476, 2020.

\bibitem[Ramasinghe and Lucey(2022)]{Ramasinghe2022Periodicity}
Sameera Ramasinghe and Simon Lucey.
\newblock Beyond periodicity: Towards a unifying framework for activations in coordinate-mlps.
\newblock In \emph{ECCV 2022: 17th European Conference Proceedings, Part XXXIII}, Berlin, Heidelberg, 2022. Springer-Verlag.
\newblock ISBN 978-3-031-19826-7.

\bibitem[Saragadam et~al.(2023)Saragadam, LeJeune, Tan, Balakrishnan, Veeraraghavan, and Baraniuk]{Saragadam2023WIREWI}
Vishwanath Saragadam, Daniel LeJeune, Jasper Tan, Guha Balakrishnan, Ashok Veeraraghavan, and Richard Baraniuk.
\newblock Wire: Wavelet implicit neural representations.
\newblock \emph{2023 IEEE/CVF Conference on Computer Vision and Pattern Recognition (CVPR)}, pages 18507--18516, 2023.

\bibitem[Rahaman et~al.(2018)Rahaman, Baratin, Arpit, Dr{\"a}xler, Lin, Hamprecht, Bengio, and Courville]{Rahaman2018OnTS}
Nasim Rahaman, Aristide Baratin, Devansh Arpit, Felix Dr{\"a}xler, Min Lin, Fred~A. Hamprecht, Yoshua Bengio, and Aaron~C. Courville.
\newblock On the spectral bias of neural networks.
\newblock In \emph{International Conference on Machine Learning}, 2018.

\bibitem[Rahimi and Recht(2007)]{Rahimi2007RandomFF}
Ali Rahimi and Benjamin Recht.
\newblock Random features for large-scale kernel machines.
\newblock In \emph{Neural Information Processing Systems}, 2007.

\bibitem[Mildenhall et~al.(2020)Mildenhall, Srinivasan, Tancik, Barron, Ramamoorthi, and Ng]{Mildenhall2020NeRFRS}
Ben Mildenhall, Pratul~P. Srinivasan, Matthew Tancik, Jonathan~T. Barron, Ravi Ramamoorthi, and Ren Ng.
\newblock Nerf: Representing scenes as neural radiance fields for view synthesis.
\newblock \emph{Commun. ACM}, 65:\penalty0 99--106, 2020.

\bibitem[Zheng et~al.(2022)Zheng, Ramasinghe, Li, and Lucey]{Zheng2021RethinkingPE}
Jianqiao Zheng, Sameera Ramasinghe, Xueqian Li, and Simon Lucey.
\newblock Trading positional complexity vs. deepness in coordinate networks.
\newblock \emph{Proceedings of the European Conference on Computer Vision (ECCV)}, 2022.

\bibitem[Wang et~al.(2021{\natexlab{c}})Wang, Liu, Yang, and Tong]{Wang2021SplinePE}
Peng-Shuai Wang, Yang Liu, Yu-Qi Yang, and Xin Tong.
\newblock Spline positional encoding for learning 3d implicit signed distance fields.
\newblock In \emph{International Joint Conference on Artificial Intelligence}, 2021{\natexlab{c}}.

\bibitem[Zeng et~al.(2024)Zeng, Burghardt, and Gambaruto]{zeng2024rbfpinnnonfourierpositionalembedding}
Chengxi Zeng, Tilo Burghardt, and Alberto~M Gambaruto.
\newblock Rbf-pinn: Non-fourier positional embedding in physics-informed neural networks, 2024.
\newblock URL \url{https://arxiv.org/abs/2402.08367}.

\bibitem[Wang et~al.(2023)Wang, Sankaran, Wang, and Perdikaris]{Wang2023AnEG}
Sifan Wang, Shyam Sankaran, Hanwen Wang, and Paris Perdikaris.
\newblock An expert's guide to training physics-informed neural networks.
\newblock \emph{ArXiv}, abs/2308.08468, 2023.

\bibitem[Cuomo et~al.(2022)Cuomo, Cola, Giampaolo, Rozza, Raissi, and Piccialli]{Cuomo2022ScientificML}
Salvatore Cuomo, Vincenzo Schiano~Di Cola, Fabio Giampaolo, Gianluigi Rozza, Maizar Raissi, and Francesco Piccialli.
\newblock Scientific machine learning through physics–informed neural networks: Where we are and what’s next.
\newblock \emph{Journal of Scientific Computing}, 92, 2022.

\bibitem[Hao et~al.(2022)Hao, Liu, Zhang, Ying, Feng, Su, and Zhu]{Hao2022PhysicsInformedML}
Zhongkai Hao, Songming Liu, Yichi Zhang, Chengyang Ying, Yao Feng, Hang Su, and Jun Zhu.
\newblock Physics-informed machine learning: A survey on problems, methods and applications.
\newblock \emph{ArXiv}, abs/2211.08064, 2022.

\bibitem[Matthews et~al.(2018)Matthews, Rowland, Hron, Turner, and Ghahramani]{Matthews2018GaussianPB}
Alexander Matthews, Mark Rowland, Jiri Hron, Richard~E. Turner, and Zoubin Ghahramani.
\newblock Gaussian process behaviour in wide deep neural networks.
\newblock 2018.

\bibitem[Long(2021)]{long2021rffpytorch}
Joshua~M. Long.
\newblock Random fourier features pytorch.
\newblock \emph{GitHub. Note: https://github.com/jmclong/random-fourier-features-pytorch}, 2021.

\end{thebibliography}
}

\newpage
\appendix

\section{Abbreviations and Notations}
\begin{table}[h]
\centering
\caption{Long forms for the abbreviations used in the paper}
\begin{tabular}{|c|c|}
\hline
Abbreviations & Long forms          \\ \hline
BC           & Boundary Condition      \\
BE           & Basic Encoding             \\
CG           & Complex Gaussian         \\
CK           & Conjugate Kernel        \\
CT           & Complex Triangle            \\
DEs          & Differential Equations  \\
FF           & Fourier Feature   \\
IC           & Initial Condition       \\
i.i.d.       & independent and identically distributed \\
L2RE         & Relative $\ell^2$ error \\
MLP          & Multi-Layer Perception  \\
NLP          & Natural Language Processing \\
NN           & Neural Network          \\
NTK          & Neural Tangent Kernel   \\
PDEs         & Partial Differential Equations  \\
PE           & Positional Encoding             \\
PIML         & Physics-Informed Machine Learning \\
PINNs        & Physics-Informed Neural Networks \\
RBF          & Radial Basis Function               \\
RBF-COM      & RBF with Compact Support            \\
RBF-INT      & RBF with Interpolants              \\
RBF-POL      & RBF with Polynomials                \\
SF           & Sinusoidal Feature                 \\ 
w.h.p.        & with high probability \\\hline
\end{tabular}
\end{table}

\begin{table}[h]
\centering
\caption{Symbols and their definitions in the paper}
\begin{tabular}{|c|c||c|c|}
\hline
Symbols & Definition  &  Symbols & Definition     \\ \hline
$a$&Activation Function& $t$&Temporal Coordinate\\
$A$&Fourier Series Coefficients&$T$&Temporal Range\\
$b$&Biases&$\mathcal{T}$&Temporal Domain\\
$\mathbf{b}$&Random sample&$u$&Differential Functions\\
$\mathcal{B}$&Boundary Operator&$\hat u_\theta$&Implicit Function\\
$\mathbf{c}$&Centres of RBFs&$w$&Weights\\
$d$&Number of Neurons&$x$&Spatial Coordinate or Input\\
$\mathcal{D}$&Differential Operator&$\mathbf{x}$&Set of input vector\\
$f$&Layer Function&$\mathcal{X}$&Matrices after mapping or activation\\
$F$&Arbitrary Function &$\bigvee$&Or\\
$G$&Arbitrary Function &$\forall$&For All\\
$h$&Hidden Layer Function&$\xi$&Cut-off distance\\
$H$&Arbitrary Function &$\varphi$&Feature Mapping Function\\
$\mathbf{K}$&Kernel&$\Phi$&Feature Space\\
$l$&Layer&$\lambda$&Loss Weighting\\
$\mathcal{L}$&Loss&$\theta$&Model Parameters\\
$N$&Number of sample points&$\Theta$&Conjugate Kernel\\
$P$&Polynomial Function&$\Sigma$&Neural Tangent Kernel\\
$r$&Distance & $\gamma$&Surjectivity Hyperparameter\\ 
$\mathbb{R}$&Real Number& $\Omega$&Spatial Domain\\ 

\hline
\end{tabular}
\end{table}

\newpage
\section{Related Work}
\textbf{Coordinate Sampling.}
As a mesh-free method, PINNs are normally evaluated on scattered collocation points both on the interior domain and IC/BC.  Therefore, the sampling strategy is crucial to PINNs' performance and efficiency. An insufficient distributed initial sampling can lead to the PDE system being ill-conditioned and NN training instability. The whole design of experiments on the fixed input sampling is reviewed by~\cite{Das2022StateoftheArtRO}. Based on the study of uniform sampling, ~\citet{WU2023115671} proposed an adaptive sampling scheme that refines high residual area during training. Similarly, Importance Sampling inspired by Monte Carlo approximation is investigated by~\cite{Nabian2021EfficientTO, Yang23DMIS}.~\citet{Daw22Propagation} proposed a novel sampling strategy that mitigates the `propagation failure' of solutions from IC/BC to the PDE residual field. Recently, \citet{Lau2024PINNACLEPA} presented a work that adaptively select collocation and experimental points.

\textbf{Novel Activation.}
The activation function in the MLP has been found to play an important role in the convergence of the PINNs. Popular activation ReLU is deficient for high-order PDEs since its second-order derivative is 0. Apart from the standard Tanh activation~\cite{Raissi2019PhysicsinformedNN}, layer-wise and neuron-wise adaptive activation are proven to be useful to accelerate the training~\cite{Jagtap2019AdaptiveAF, Jagtap2020LocallyAA}. Another line of seminal work, SIREN~\cite{Sitzmann2020ImplicitNR}, which uses periodic activation function, has achieved remarkable results in  Neural Representation and tested on solving the Poisson equation. Gaussian~\cite{Ramasinghe2022Periodicity} and Gabor Wavelet activations~\cite{, Saragadam2023WIREWI} are proven to be effective alternatives.

\textbf{Positional Embedding.} Broadly speaking, PINNs can also be considered as a special type of Neural Fields~\cite{Xie2021NeuralFI} in visual computing, which specifically feed coordinate-based input to MLPs that represent continuous field quantity (e.g. velocity field in fluid mechanics) over arbitrary spatial and temporal resolution. However, the PINNs community has largely overlooked that both perspectives function the same way as Implicit Neural Representations. In the Neural Field, images and 3D shapes are naturally high-frequency signals, whereas deep networks are inherently learning towards the low-frequency components~\cite{Rahaman2018OnTS}. Feature mapping has hence become a standard process in practice that maps the low-dimension coordinates to high-dimension space. The pioneering work was conducted by~\cite{Rahimi2007RandomFF}, who used Fourier features to approximate any stationary kernel principled by Bochner's theorem. the derivative works are done such as Positional Encoding~\cite{Mildenhall2020NeRFRS}, Random Feature~\cite{Tancik2020FourierFL} and Sinusoidal Feature~\cite{Sitzmann2020ImplicitNR}. Another concurrent work discusses non-periodic feature mapping~\cite{Zheng2021RethinkingPE, Ramasinghe2021ALR, Wang2021SplinePE, zeng2024rbfpinnnonfourierpositionalembedding}. To the best of our knowledge, feature mapping in PINNs has not been comprehensively investigated. Only a few work carry out preliminary study adopting Fourier-feature-based methods in PINNs~\cite{WANG2021113938, Wang2023AnEG, Wong2022Sinusoidal}. 

For further reviews on PINNs, we refer the readers to~\cite{Cuomo2022ScientificML, Hao2022PhysicsInformedML}.

\newpage
\section{Prior Theories}
\subsection{Spectral Bias in PINNs~\cite{Wang2020WhenAW}}
\label{Append: Spectral}
Normally PINNs are setup as a standard MLP model $f(\mathbf{x};\theta)$, and $\theta$ is optimized on the loss function $L(\theta) = \frac{1}{2}\left |f(\mathbf{x}; \theta) - Y\right|^2= \frac{1}{2}\sum_{i}^{N}(f(x_i; \theta)-y_i)^2$, where $X$, $Y$ and $\theta$ are training input, training ground truth and model parameters. For an easier formulation, we replace the conventional gradient descent formulation $\theta_{t+1} = \theta_t - \alpha \nabla_{\theta}L(\theta_t)$ to a gradient flow equation:
\begin{equation}
\frac{d\theta}{dt} = - \alpha \nabla_{\theta}L(\theta_t),
\end{equation}
where $\alpha$ should be an infinitesimally small learning rate in the NTK setting. 
Given PDE collocation data points$\{x_r^i, \mathcal{D}(\hat u_\theta(x_{r}^i)) \}_{i = 1}^{N_r}$, and boundary training points$\{x_{bc}^i, \mathcal{B}(\hat u_\theta(x_{bc}^i))\}_{i = 1}^{N_{b}}$. The gradient flow can be formulated as~\cite{Wang2020WhenAW}:
\begin{equation}
\label{equa: pinn gradient flow}
\left[\begin{array}{c}
\frac{d u\left(x_b, \theta_t\right)}{d t} \\
\frac{d \mathcal{L} u\left(x_r, \theta_t\right)}{d t}
\end{array}\right] = -\left[\begin{array}{ll}
\boldsymbol{K}_{u u}^t & \boldsymbol{K}_{u r}^t \\
\boldsymbol{K}_{r u}^t & \boldsymbol{K}_{r r}^t
\end{array}\right] \cdot \left[\begin{array}{c}
u\left(x_b, \theta_t\right)- \mathcal{B}(\hat u_\theta(x_{b})) \\
\mathcal{L} u\left(x_r, \theta_t\right)-\mathcal{D}(\hat u_\theta(x_{r}))
\end{array}\right],
\end{equation}
where the Kernels $\boldsymbol{K}$ are:
\begin{equation}
\begin{aligned}
& \left(\boldsymbol{K}_{u u}^t\right)_{i j}=\left\langle\frac{d u\left(x_b^i, \theta_t\right)}{d \theta}, \frac{d u\left(x_b^j, \theta_t\right)}{d \theta}\right\rangle, \\
& \left(\boldsymbol{K}_{r r}^t\right)_{i j}=\left\langle\frac{d \mathcal{L}\left(x_r^i, \theta_t\right)}{d \theta}, \frac{d \mathcal{L}\left(x_r^j, \theta_t\right)}{d \theta}\right\rangle,\\
& \left(\boldsymbol{K}_{u r}^t\right)_{i j}=\left(\boldsymbol{K}_{r u}^t\right)_{i j}=\left\langle\frac{d u\left(x_b^i, \theta_t\right)}{d \theta}, \frac{d \mathcal{L} u\left(x_r^j, \theta_t\right)}{d \theta}\right\rangle,
\end{aligned}
\end{equation}
Since $\boldsymbol{K}$ remains stationary, then $\boldsymbol{K}^t \approx \boldsymbol{K}^0$ as NN width tends to infinity, Equation~\ref{equa: pinn gradient flow} is rewritten as:
\begin{equation}
\begin{aligned}
\left[\begin{array}{c}
\frac{d u\left(x_b, \theta_t\right)}{d t} \\
\frac{d \mathcal{L} u\left(x_r, \theta_t\right)}{d t}
\end{array}\right] & \approx - \boldsymbol{K}^0 \left[\begin{array}{c}
u\left(x_b, \theta_t\right)- \mathcal{B}(\hat u_\theta(x_{b})) \\
\mathcal{L} u\left(x_r, \theta_t\right)-\mathcal{D}(\hat u_\theta(x_{r}))
\end{array}\right], \\
& \approx(I - e^{-\boldsymbol{K}^0t}) \cdot\left[\begin{array}{l}
\mathcal{B}(\hat u_\theta(x_{b}) \\
\mathcal{D}(\hat u_\theta(x_{r}))
\end{array}\right],
\end{aligned}
\end{equation}
By Schur product theorem, $\boldsymbol{K}^0$ is always Positive Semi-definite, hence it can be Eigen-decomposed to $\boldsymbol{Q}^T \Lambda \boldsymbol{Q}$, where $\boldsymbol{Q}$ is an orthogonal matrix and $\Lambda$ is a diagonal matrix with eigenvalues $\lambda _i$ in the entries. We can rearrange the training error in the form of:
\begin{equation}
\begin{aligned}
\left[\begin{array}{c}
\frac{d u\left(x_b, \theta_t\right)}{d t} \\
\frac{d \mathcal{L} u\left(x_r, \theta_t\right)}{d t}
\end{array}\right] - \left[\begin{array}{l}
\mathcal{B}(\hat u_\theta(x_{b}) \\
\mathcal{D}(\hat u_\theta(x_{r}))
\end{array}\right] & \approx (I - e^{-\boldsymbol{K}^0t}) \cdot \left[\begin{array}{l}
\mathcal{B}(\hat u_\theta(x_{b}) \\
\mathcal{D}(\hat u_\theta(x_{r}))
\end{array}\right] - \left[\begin{array}{l}
\mathcal{B}(\hat u_\theta(x_{b}) \\
\mathcal{D}(\hat u_\theta(x_{r}))
\end{array}\right], \\
& \approx -\boldsymbol{Q}^T e^{-\Lambda t} \boldsymbol{Q} \cdot \left[\begin{array}{l}
\mathcal{B}(\hat u_\theta(x_{b}) \\
\mathcal{D}(\hat u_\theta(x_{r}))
\end{array}\right], \\
\end{aligned}
\end{equation}
where $e^{-\Lambda t} = \left[\begin{array}{cccc} e^{-\lambda_1 t} & & \\
& \ddots & \\
& & e^{-\lambda_N t}
\end{array}\right] $. This indicates the decrease of training error in each component is exponentially proportional to the eigenvalues of the deterministic NTK, and the NN is inherently biased to learn along larger eigenvalues entries of the $\boldsymbol{K}^0$.
\subsection{Input Gradient Variability~\cite{Wong2022Sinusoidal}}
\label{Append: Input Gradient Variability}
The input gradient for arbitrary input x can be derived using the chain rule:
\begin{equation}
\begin{aligned}
& \frac{\partial \hat u_\theta}{\partial x}=\sum_{j=1}^n w_{l, j} a^{\prime}\left(u_{l-1, j}\right) \frac{\partial u_{l-1, j}}{\partial x}, \\
& \frac{\partial u_{l j}}{\partial x}=\sum_{i=1}^n w_{l, i j} a^{\prime}\left(u_{l-1, i}\right) \frac{\partial u_{l-1, i}}{\partial x} \text {, for } 1<l<L, \\
& \frac{\partial u_{1, i}}{\partial x}=w_{1, i},
\end{aligned}
\end{equation}
where $x_{l, j} = a(u_{l,j})$ is the $j$-th input of the $l$-th hidden layer. \\
Following, we give an example of how to derive the input gradient distribution at initialisation.\\
Let $\hat u(x; w)$ be the PINN with $L$ fully connected layers and $n$ neurons in eahc layer, activation function $a = tanh$ and parameters $W$. The network is initialised by Xavier initialisation.
The mean of the input gradient is given by:
\begin{equation}
\mathbf{E}\left[\frac{\partial \hat{u}}{\partial x}\right]=n \mathbf{E}\left[w_l a^{\prime}\left(u_{l-1}\right) \frac{\partial u_{l-1}}{\partial x}\right]=n \mathbf{E}\left[w_l\right] \mathbf{E}\left[a^{\prime}\left(u_{l-1}\right) \frac{\partial u_{l-1}}{\partial x}\right]=0,
\end{equation}
The variance of the input gradient is given by:
\begin{equation}
\begin{aligned}
\mathbf{Var}\left(\frac{\partial \hat{u}}{\partial x}\right) &=n \mathbf{Var}\left(w_l a^{\prime}\left(u_{l-1}\right) \frac{\partial u_{l-1}}{\partial x}\right), \\ 
&= n \mathbf{Var}\left(w_L\right) \mathrm{E}\left[\left(a^{\prime}\left(u_{l-1}\right) \frac{\partial u_{l-1}}{\partial x}\right)^2\right] \leq n \mathbf{Var}\left(w_l\right) \mathbf{Var}\left(\frac{\partial u_{l-1}}{\partial x}\right), \\ 
&= \frac{2 n}{n+1} \mathbf{Var}\left(\frac{\partial u_{l-1}}{\partial x}\right),
\end{aligned}
\end{equation}
Since $a'=sech^2$, for any layer $1\le l < L$, $0< a(u_l) \le 1$.

\begin{equation}
\mathbf{Var}\left(\frac{\partial u_l}{\partial x}\right) \leq n \mathbf{Var}\left(w_l\right) \mathbf{Var}\left(\frac{\partial u_{l-1}}{\partial x}\right)=\mathbf{Var}\left(\frac{\partial u_{l-1}}{\partial x}\right),
\end{equation}
Under Xavier Initialisation, $\textbf{Var}\left(\frac{\partial u_l}{\partial x}\right) = \frac{2}{n+1}$. We can get:

\begin{equation}
\begin{aligned}
& \mathbf{Var}\left(\frac{\partial \hat{u}}{\partial x}\right) \leq \frac{2 n}{n+1} \mathbf{Var}\left(\frac{\partial u_{l-1}}{\partial x}\right) \leq \frac{2 n}{n+1} \mathbf{Var}\left(\frac{\partial u_{l-2}}{\partial x}\right) \leq \cdots \leq \\
& \frac{2 n}{n+1} \mathbf{Var}\left(\frac{\partial u_2}{\partial x}\right) \leq \frac{2 n}{n+1} \mathbf{Var}\left(\frac{\partial u_1}{\partial x}\right)=\frac{2 n}{n+1} \frac{2}{n+1}
\end{aligned}
\end{equation}
This reveals the variance of the input gradient tends to $0$ as the width of the layers tend to be infinite. Zero input gradient leads to a constant output and higher derivatives $\frac{\partial^2 \hat{u}}{\partial x^2}, \frac{\partial^3 \hat{u}}{\partial x^3}, \ldots, \frac{\partial^k \hat{u}}{\partial x^k}$ are $0$. And ultimately, the surrogate model for the differential equations $\mathcal{D}(\frac{\partial^2 \hat{u}}{\partial x^2}, \frac{\partial^3 \hat{u}}{\partial x^3}, \ldots, \frac{\partial^k \hat{u}}{\partial x^k}) = 0$. This suggests the PINNs with wide layers can have near zero input gradient and can easily fall into local minimum at initialisation. However, the joint loss of the PDE and the BC can still be far away from the true solution. Examples of other activation functions are demonstrated in ~\cite{Wong2022Sinusoidal}.

\section{Proof of Lemma \ref{lemma: inject}}
\label{Append: inject}
\textbf{Lemma \ref{lemma: inject}}. Consider a randomly sampled and normalised input $\mathbf{x}=[x_1, x_2, \cdots, x_n]^T$, $x \in [0, 1]^d$, and its corresponding features in $\Phi: \mathbb{R}^d \rightarrow \mathbb{R}^m = [\varphi(x_1), \varphi(x_2), \cdots, \varphi(x_n)]^T$, let the feature mapping function $\varphi(\mathbf{x}) = sin(2\pi \mathcal{B} \mathbf{x}) \in [-1, 1]$, where $\mathcal{B}$ is sampled from a Gaussian distribution $\mathcal{N}(0, \sigma)$, the mapping function $sin(\cdot)$ is surjective w.h.p.\\
Proof. Since $\mathbf{x} \in [0, 1]$, then $\Phi(\mathbf{x}) \in [sin(0), sin(2\pi \mathcal{B})]$. Noting that $sin(\cdot)$ is only bijective on $(-\pi,\pi)$ or $sin(\cdot)$ is bijective on $[0, 2\pi)$ only if $x \neq \pi$ within the domain limit. Hence we can derive the probability $\mathbb{P}$ of $sin(\cdot)$ is a bijection in following inequality form:\\
\begin{equation}
\begin{aligned}
\mathbb{P} &< P(0\le \mathcal{B}<1), \\
            & <P(\mathcal{B}<1) - P(\mathcal{B} < 0),\\
            &<\int_{-\infty}^{\mathbf{x}=1} \frac{1}{ \sqrt{2\pi \sigma^2}}e^{-\frac{(\mathbf{x}-\mu)^2}{2\sigma^2}} \,d\mathbf{x}\ - \int_{-\infty}^{\mathbf{x}=0} \frac{1}{ \sqrt{2\pi \sigma^2}}e^{-\frac{(\mathbf{x}-\mu)^2}{2\sigma^2}} \,d\mathbf{x}\ ,\\
            &< \frac{1}{ \sqrt{2\pi \sigma^2}} \int_{-\infty}^{\mathbf{x}=1} e^{-\frac{(\mathbf{x}-\mu)^2}{2\sigma^2}} \,d\mathbf{x}\ - \frac{1}{ \sqrt{2\pi \sigma^2}} \int_{-\infty}^{\mathbf{x}=0} e^{-\frac{(\mathbf{x}-\mu)^2}{2\sigma^2}} \,d\mathbf{x}\ ,\\
            &  \qquad
                {\textup{Substitution}}
               \quad
               \boxed{\begin{aligned}
                           z&= \frac{\mathbf{x}-\mu}{\sqrt{2}\sigma}
                      \end{aligned}} \\
            &< \frac{1}{ \sqrt{2\pi \sigma^2}}\sqrt{2}\sigma \int_{-\infty}^{\mathbf{x}=1} e^{-z^2} \,dz\ - \frac{1}{ \sqrt{2\pi \sigma^2}}\sqrt{2}\sigma \int_{-\infty}^{\mathbf{x}=0} e^{-z^2} \,dz\ , \\
            &< \frac{1}{ \sqrt{2\pi \sigma^2}}\sqrt{2}\sigma \frac{\sqrt{\pi}}{2}erf(\frac{1}{\sqrt{2}\sigma}) - \frac{1}{ \sqrt{2\pi \sigma^2}}\sqrt{2}\sigma \frac{\sqrt{\pi}}{2}erf(0), \\
            & < \frac{1}{2}erf(\frac{1}{\sqrt{2}\sigma}),
\end{aligned}
\end{equation}
We calculated the upper bound of $\mathbb{P}$ for the $sin(\cdot)$ to be bijective is less than 0.5 as $\sigma \rightarrow 0$, and decreases as $\sigma$ increases. Hence $sin(\cdot)$ is surjective w.h.p is proved by contrapositive.

\section{Proof of Proposition \ref{Prop: ck}} 
\label{Append: ck}
\textbf{Proposition \ref{Prop: ck}} (Propagation of the Conjugate Kernel).
Let input $x\in \mathbb{R}^{N\times n}$, and each layer of the Neural Network is parameterised with independent and identically distributed (i.i.d.) weights and biases from standard Gaussian distribution. Hence $f^l(\mathbf{x}; \theta_0)\sim \mathcal{GP}(0, \Sigma^{l}(\mathbf{x}, \mathbf{x}'))$, and the Conjugate Kernels propagate through the Neural Network in the following recursive form:\\
\begin{equation}
\begin{aligned}
 \Sigma^{0}(\mathbf{x}, \mathbf{x}')&= \langle \mathbf{x}, \mathbf{x}'\rangle + 1,\\
 \Sigma^1(\mathbf{x}, \mathbf{x}')&= \mathbf{E}[\varphi(\mathbf{x})^T\varphi(\mathbf{x}')] + 1,\\
 \Sigma^l(\mathbf{x}, \mathbf{x}')&= \mathbf{E}[a(\mathcal{X})^Ta(\mathcal{X}')] + 1, \quad 2 \leq l \leq L,\\
\end{aligned}
\end{equation}
where $\varphi$ is the feature mapping function at $l=1$ and $\mathcal{X}, \mathcal{X}'$ are the hidden layer state from previous layer and $\left[\begin{array}{cc} \mathcal{X} \\ \mathcal{X}' \end{array}\right]\sim \mathcal{N}  (\left[\begin{array}{cc} 0 \\0 \end{array}\right], \left[\begin{array}{cc}\Sigma^{l-1}(\mathbf{x}, \mathbf{x}) & \Sigma^{l-1}\left(\mathbf{x}^{\prime}, \mathbf{x}\right) \\ \Sigma^{l-1}\left(\mathbf{x}, \mathbf{x}^{\prime}\right) & \Sigma^{l-1}\left(\mathbf{x}^{\prime}, \mathbf{x}^{\prime}\right)\end{array}\right])$.\\

Proof. 
\begin{remark}
$X\sim \mathcal{N}(\mu_\mathbf{x}, \Sigma_\mathbf{x})$ is equivalent to $X\sim \mu_\mathbf{x} + \Sigma_\mathbf{x}\mathcal{N}(0, 1)$. Hence if $Y = a + bX$, then $Y\sim \mathcal{N}(a+ b\mu_\mathbf{x}, b\Sigma_\mathbf{x}b^T)$.
\end{remark}
Recall Equation~\ref{Equa: MLP} and~\ref{Equa: feature mapping}, the values of $f^{l=2}$ depend on the post feature mapping layer and the values of $f^{2\leq l\leq L}$ depend on the previous layer. We treat each $f(\mathbf{x};\theta) = \frac{1}{n^l} \phi(\mathbf{x})\theta$, where $\phi(\mathbf{x})$ can represent the feature mapping function $\varphi(\mathbf{x})$ or activation function $a(\mathbf{x})$. Since $\theta \sim \mathcal{N}(0, 1)$, then $\mathbf{Var}[f(\mathbf{x};\theta)] = \phi(\mathbf{x})^T\phi(\mathbf{x}) = K(\mathbf{x}, \mathbf{x}')$, as $w\phi(\mathbf{x})+b$ is a linear transformation. 

Then the layers can be described as a Gaussian Process with mean 0 and covariance $K(\mathbf{x}, \mathbf{x}') = \phi(\mathbf{x})^T\phi(\mathbf{x}) = 
\begin{cases}
\frac{1}{n^{l-1}}\varphi{(f^{l-1}(\mathbf{x}; \theta_0))}^T\varphi{(f^{l-1}(\mathbf{x}; \theta_0))}, & l=1,\\
\frac{1}{n^{l-1}}a{(f^{l-1}(\mathbf{x}; \theta_0))}^Ta{(f^{l-1}(\mathbf{x}; \theta_0))}, & 2\leq l\leq L,\\
\end{cases}$ \\

The vector form $f$ is a summation of its each components: $\frac{1}{n^{l-1}}\Phi(f(\mathbf{x}; \theta_0))^T\Phi(f(\mathbf{x}'; \theta_0)) = \frac{1}{n^{l-1}}\sum_{i=1}^{n^{l-1}}\Phi(f_i(\mathbf{x}; \theta_0))^T\Phi(f_i(\mathbf{x}'; \theta_0))$.
Since the $f$s are independent, by applying the Law of Large Numbers, similarly in~\cite{Matthews2018GaussianPB}, we can get $\frac{1}{n^{l-1}}\phi(f(\mathbf{x}; \theta_0))^T\phi(f(\mathbf{x}'; \theta_0)) = \mathbf{E}[\phi(f(\mathbf{x}; \theta_0))^T\phi(f(\mathbf{x}'; \theta_0))] = \mathbf{E}[\phi(\mathcal{X})^T\phi(\mathcal{X}')]$ where $\mathbf{Cov}(\mathcal{X},\mathcal{X}') = \left[\begin{array}{cc}\Sigma^{l-1}(\mathbf{x}, \mathbf{x}) & \Sigma^{l-1}\left(\mathbf{x}^{\prime}, \mathbf{x}\right) \\ \Sigma^{l-1}\left(\mathbf{x}, \mathbf{x}^{\prime}\right) & \Sigma^{l-1}\left(\mathbf{x}^{\prime}, \mathbf{x}^{\prime}\right)\end{array}\right]$.

\section{Proof of Theorem \ref{theorem: ntk}} 
\label{Append: ntk}
\textbf{Theorem \ref{theorem: ntk}} (Evolution of the NTK with CK).
Let input $\mathbf{x}\in \mathbb{R}^{N\times n}$, $\phi(\mathbf{x})=\varphi(\mathbf{x}) \bigvee a(\mathbf{x})$ ; Recall $\Sigma^1(\mathbf{x}, \mathbf{x}')= \mathbf{E}[\varphi(\mathbf{x})^T\varphi(\mathbf{x}')] + 1$, $\Sigma^l(\mathbf{x},\mathbf{x}') = \mathbf{E}[\phi(\mathbf{x})\phi(\mathbf{x}')]+1$ and its derivative is $\dot{\Sigma}^l(\mathbf{x},\mathbf{x}') = \mathbf{E}[\dot{\phi}(\mathbf{x})\dot{\phi}(\mathbf{x}')], \in  \mathbb{R}^{N \times N}$. Assuming the infinity width limit, the gradient $\nabla f^l$ satisfies:\\
\begin{equation}
\nabla_\theta f^l(\mathbf{x}; \theta_0)^T\nabla_\theta f^l(\mathbf{x}'; \theta_0) \to \Theta^l(\mathbf{x}, \mathbf{x}'),
\end{equation}
The evolution of the kernels follows:
\begin{equation}
\begin{aligned}
\Theta^1(\mathbf{x}, \mathbf{x}') &= \Theta^{0}(\mathbf{x}, \mathbf{x}')\dot{\Sigma}^1(\mathbf{x}, \mathbf{x}') + \Sigma^1(\mathbf{x}, \mathbf{x}'),\\
\Theta^l(\mathbf{x}, \mathbf{x}') &= \Theta^{l-1}(\mathbf{x}, \mathbf{x}')\dot{\Sigma}^l(\mathbf{x}, \mathbf{x}') + \Sigma^l(\mathbf{x}, \mathbf{x}'), \quad 2 \leq l \leq L,
\end{aligned}
\end{equation}
Proof. Since the NTK involves derivative with the $\theta$, we need to consider the $\theta$s in both the previous layer and the current layer, Thus we formulate $\theta^l = \theta^{l-1} \cup \theta^{l*} = \theta^{l-1} \cup \{w^l, b^l\}$, which gives $f^l(\mathbf{x}; \theta^l) = \frac{1}{\sqrt{n^{l-1}}}w^l\phi(f^{l-1}(\mathbf{x};\theta^{l-1})) + b^l$. With the new notation, we can split the derivatives by partial differention rules:
\begin{equation}
\begin{aligned}
\nabla_{\theta^l}f^l(\mathbf{x};\theta^l)^T\nabla_{\theta^l}f^l(\mathbf{x};\theta^l)&=\nabla_{\theta^{l*}}f^l(\mathbf{x};\theta^l)^T\nabla_{\theta^{l*}}f^l(\mathbf{x};\theta^l) + \nabla_{\theta^{l-1}}f^l(\mathbf{x};\theta^l)^T\nabla_{\theta^{l-1}}f^l(\mathbf{x};\theta^l),\\
&=\frac{1}{n^{l-1}}\phi(f^{l-1}(\mathbf{x};\theta^{l-1}))^T\phi(f^{l-1}(\mathbf{x};\theta^{l-1})) + \nabla_{\theta^{l-1}}f^l(\mathbf{x};\theta^l)^T\nabla_{\theta^{l-1}}f^l(\mathbf{x};\theta^l),\\
&=\begin{cases}
\Sigma^2(\mathbf{x},\mathbf{x}')\\
\Sigma^l(\mathbf{x},\mathbf{x}'), & 2 \leq l \leq L\\
\end{cases} + \nabla_{\theta^{l-1}}f^l(\mathbf{x};\theta^l)^T\nabla_{\theta^{l-1}}f^l(\mathbf{x};\theta^l),\\
\end{aligned}
\end{equation}

The first part of the partial differention becomes precisely the Conjugate Kernel that is derived from Proposition~\ref{Prop: ck}. The remaining part can be solved by chain rule.\\

\begin{equation}
\begin{aligned}
\nabla_{\theta^{l-1}}f^l(\mathbf{x};\theta^l)^T\nabla_{\theta^{l-1}}f^l(\mathbf{x};\theta^l) &= \frac{1}{\sqrt{n^{l-1}}}w^l\nabla_{\theta^{l-1}}\phi(f^{l-1}(\mathbf{x};\theta^{l-1}))^T\frac{1}{\sqrt{n^{l-1}}}w^l\nabla_{\theta^{l-1}}\phi(f^{l-1}(\mathbf{x};\theta^{l-1})),\\
&= \frac{1}{\sqrt{n^{l-1}}}w^l diag[\phi'(f^{l-1}(\mathbf{x};\theta^{l-1}))]\nabla_{\theta^{l-1}}(f^{l-1}(\mathbf{x};\theta^{l-1}))^T,\\&\frac{1}{\sqrt{n^{l-1}}}w^ldiag[\phi'(f^{l-1}(\mathbf{x};\theta^{l-1}))]\nabla_{\theta^{l-1}}(f^{l-1}(\mathbf{x};\theta^{l-1})),\\
&= \frac{1}{n^{l-1}}w^l diag[\phi'(f^{l-1}(\mathbf{x};\theta^{l-1}))] \underbrace{(\nabla_{\theta^{l-1}}(f^{l-1}(\mathbf{x};\theta^{l-1}))^T\nabla_{\theta^{l-1}}(f^{l-1}(\mathbf{x};\theta^{l-1})))}_\text{NTK}\\&diag[\phi'(f^{l-1}(\mathbf{x};\theta^{l-1}))]{w^l}^{T},\\
&= \frac{1}{n^{l-1}}\sum_i^{n^{l-1}} w^l_i \phi'(f^{l-1}(\mathbf{x};\theta^{l-1}))\Theta^{l-1}\phi'(f^{l-1}(\mathbf{x};\theta^{l-1})){w^l_i}^{T},\\
&= \Theta^{l-1}\frac{1}{n^{l-1}}\sum_i^{n^{l-1}} w^l_i \underbrace{\phi'(f^{l-1}(\mathbf{x};\theta^{l-1}))\phi'(f^{l-1}(\mathbf{x};\theta^{l-1}))}_\text{derivative of CK}{w^l_i}^{T},\\
&= \begin{cases}
\Theta^1(\mathbf{x}, \mathbf{x}') = \Theta^{0}(\mathbf{x}, \mathbf{x}')\dot{\Sigma}^1(\mathbf{x}, \mathbf{x}'),\\
\Theta^l(\mathbf{x}, \mathbf{x}') = \Theta^{l-1}(\mathbf{x}, \mathbf{x}')\dot{\Sigma}^l(\mathbf{x}, \mathbf{x}'), 2\leq l\leq L,\\
\end{cases}
\end{aligned}
\end{equation}

\newpage

\section{Different feature mapping methods in MLP}
\label{Append: FM}
\textbf{Basic Encoding:}~\cite{Mildenhall2020NeRFRS} $\varphi(\mathbf{x}) = [cos(2\pi\mathbf{x}), sin(2\pi\mathbf{x})]^T$ for $j = 0, .., m-1$.\\
\textbf{Positional Encoding:}~\cite{Mildenhall2020NeRFRS} $\varphi(\mathbf{x}) = [cos(2\pi\sigma^{j/m}\mathbf{x}), sin(2\pi\sigma^{j/m}\mathbf{x})]^T$ for $j = 0, .., m-1$.\\
\textbf{Random Fourier:}~\cite{Tancik2020FourierFL} $\varphi(\mathbf{x}) = [cos(2\pi \sigma \mathcal{B} \mathbf{x}), sin(2\pi \sigma \mathcal{B} \mathbf{x})]^T$, where $\mathcal{B} \in \mathbb{R}^{m\times d}$ is sampled from $\mathcal{N}(0, 1)$ and $\sigma$ is an arbitrary scaling factor varies case to case.\\
\textbf{Sinusoidal  Feature:}~\cite{Sitzmann2020ImplicitNR} $\varphi(\mathbf{x}) = [sin(2\pi \mathbf{W} \mathbf{x} + \mathbf{b})]^T$, where $\mathbf{W}$ and $\mathbf{b}$ are trainable parameters.\\
\textbf{Complex Triangle:}~\cite{Zheng2021RethinkingPE} $\varphi(\mathbf{x}) = [max(1-\frac{|x_1-t|}{0.5d}, 0), max(1-\frac{|x_2-t|}{0.5d}, 0), \cdots, max(1-\frac{|x_i-t|}{0.5d}, 0)]^T$, where $t$ is uniformly sampled from 0 to 1.\\
\textbf{Complex Gaussian:}~\cite{Zheng2021RethinkingPE} $\varphi(\mathbf{x}) = [e^{-0.5(x_1 - \tau/d)^2/\sigma^2}\bigotimes\cdots\bigotimes e^{-0.5(x_d - \tau/d)^2/\sigma^2}]^T$, where $\tau$ is uniformly sampled from $[0, 1]$, and $\bigotimes$ is the Kronecker product.

\subsection{Example of Composed NTK using Fourier Features}
\label{Append: CNTK}
The Fourier feature layer is defined as:\\
\begin{equation}
\varphi(\mathbf{x})=\left[a_1 \cos \left(2 \pi b_1^{\mathrm{T}} \mathbf{x}\right), a_1 \sin \left(2 \pi \mathbf{b}_1^{\mathrm{T}} \mathbf{x}\right), \ldots, a_m \cos \left(2 \pi \mathbf{b}_m^{\mathrm{T}} \mathbf{x}\right), a_m \sin \left(2 \pi \mathbf{b}_m^{\mathrm{T}} \mathbf{x}\right)\right]^{\mathrm{T}},
\end{equation}\\
Hence the NTK is computed by:\\
\begin{equation}
\begin{aligned}
\boldsymbol{K}_{\Phi}\left(x_i, x_j\right)  & = \varphi(x_i)^T\varphi(x_j), \\
& =\left[\begin{array}{c}
A_k\cos \left(2\pi \boldsymbol{b_m} x_i\right) \\
A_k\sin \left(2\pi \boldsymbol{b_m} x_j\right)
\end{array}\right]^{\mathrm{T}} \cdot\left[\begin{array}{c}
A_k\cos \left(2\pi \boldsymbol{b_m} x_i\right) \\
A_k\sin \left(2\pi \boldsymbol{b_m} x_j\right)
\end{array}\right], \\
& =  \sum_{k=1}^m A_k \cos \left(2\pi b_k^T x_i\right) \cos \left(2\pi b_k^T x_j\right), \\
&+A_k \sin \left(2\pi b_k^T x_i\right) \sin \left(2\pi b_k^T x_j\right), \\
&\boxed{\begin{aligned}
                           \text{Trigonometric Identities:} \cos(c - d)&= \cos c\cos d+\sin c \sin d
                      \end{aligned}} \\
& =  \sum_{k=1}^m A_k^2 \cos \left(2\pi b_k^T\left(x_i-x_j\right)\right) ,
\end{aligned}
\end{equation}\\
where $A$ is the Fourier Series coefficients, $ \boldsymbol{b} $ is randomly sampled from $\mathcal{N}(0,\,\sigma^{2})$ and $\sigma$ is an arbitrary hyperparameter that controls the bandwidth.
Thereafter, the feature space becomes the input of the NTK which gives the identities: $\boldsymbol{K}_{NTK}(x_i^Tx_j) = \boldsymbol{K}_{NTK}(\varphi(x_i)^T\varphi(x_j)) = \boldsymbol{K}_{NTK}(\boldsymbol{K}_{\Phi}(x_i - x_j))$.

\newpage

\section{Impact Statement}
\label{appendix: impact}
The aim of this research is to contribute to the development of Machine Learning. Our work may have various implications for society, but we do not think any of them need special attention here. Although significant progress has been made in the study of PINNs, we suggest using them cautiously in real-life applications.

\section{Reproducibility}
\label{appendix: reproducibility}
\textbf{Implementation details.} All feature mapping methods are implemented in the same NN architecture that consists of 5 fully connected layers with 100 neurons each for the forward problems and 50 neurons each for the inverse problems unless otherwise specified. The numbers of features from each feature function tested are 64, 128 and 256. The numbers of polynomial terms tested are 5, 10, 15 and 20. The non-linear activation function chosen is Tanh. The NN parameters are initialised with Xavier initialisation. The NN is trained with the Adam optimiser with an initial learning rate of $1e-3$ for 20k epochs and L-BFGS for another 20k epochs. 

\textbf{Software \& Hardware.} All codes are implemented in Pytorch 2.0.0 and can be found in this [Anonymous link]. Compared feature mapping methods are implemented in public library including \textit{random-fourier-features-pytorch}~\cite{long2021rffpytorch}, \textit{siren-pytorch}~\cite{Sitzmann2020ImplicitNR} and code repository from ~\cite{Zheng2021RethinkingPE}. All codes are under MIT license. The GPUs used to carried out experiments include an Nvidia Tesla V100 PCle 16GB and an Nvidia RTX 3090 24GB.

\textbf{Evaluation.} We employed the standard mean square error (MSE) as the loss function for the PDE loss term and IC/BC loss terms, they generally have good behaviour during training. The prediction results are evaluated by a relative $\ell^2$ error.
\begin{equation}
\mathrm{L} 2 \mathrm{RE}=\sqrt{\frac{\sum_{i=1}^n\left(u_i-u_i^{\prime}\right)^2}{\sum_{i=1}^n u_i^{\prime 2}}},
\end{equation}
where $\mathbf{u}$ is the prediction results in all dimensions and $\mathbf{u}^{\prime}$ is the ground truth from either analytical solution or high-fidelity numerical methods. For the inverse problems, the $\ell^2$ is computed between the predicted coefficients and true coefficients that are used to generate the data.

\newpage

\section{Complete experimental results for Table~\ref{tab: forward pdes}\&\ref{tab: inverse}}
\label{Append: full results}
\subsection{Complete results for Table~\ref{tab: forward pdes}}

\begin{table}[h]
\caption{Full PDEs benchmark results comparing different feature mapping methods in $\ell^2$ error.  The best results are in \colorbox{blue!25}{Blue}. Standard deviations are shown after $\pm$.}
\label{tab: full forward pdes}
\centering
\begin{adjustbox}{width=\columnwidth,center}
\begin{tabular}{@{}cccccc@{}}
\toprule
                               & vanilla-PINN                           & BE                 & PE            & FF                & SF               \\ \midrule
\multicolumn{1}{l|}{Wave}      & \multicolumn{1}{l|}{3.731e-1$\pm$2.369e-2} & \multicolumn{1}{l|}{1.035e0$\pm$3.548e-1}  & \multicolumn{1}{l|}{1.014e0$\pm$4.019e-1}  & \multicolumn{1}{l|}{\colorbox{blue!25}{2.375e-3$\pm$3.751e-4}} & \multicolumn{1}{l|}{7.932e-3$\pm$9.321e-4}  \\
\multicolumn{1}{l|}{Diffusion} & \multicolumn{1}{l|}{1.426e-4$\pm$4.841e-5} & \multicolumn{1}{l|}{1.575e-1$\pm$6.128e-2} & \multicolumn{1}{l|}{1.595e-1$\pm$1.204e-2} & \multicolumn{1}{l|}{2.334e-3$\pm$7.514e-4} & \multicolumn{1}{l|}{3.474e-4$\pm$6.107e-5} \\
\multicolumn{1}{l|}{Heat}      & \multicolumn{1}{l|}{4.732e-3$\pm$6.140e-5} & \multicolumn{1}{l|}{6.491e-3$\pm$6.365e-4} & \multicolumn{1}{l|}{7.574e-3$\pm$1.025e-4} & \multicolumn{1}{l|}{2.190e-3$\pm$3.125e-4} & \multicolumn{1}{l|}{3.961e-3$\pm$2.568e-4} \\
\multicolumn{1}{l|}{Poisson}   & \multicolumn{1}{l|}{3.618e-3$\pm$1.236e-4} & \multicolumn{1}{l|}{4.964e-1$\pm$2.146e-2} & \multicolumn{1}{l|}{4.910e-1$\pm$1.084e-2} & \multicolumn{1}{l|}{7.586e-4$\pm$9.013e-5}  & \multicolumn{1}{l|}{9.078e-4$\pm$1.024e-5} \\
\multicolumn{1}{l|}{Burgers'}   & \multicolumn{1}{l|}{1.864e-3$\pm$1.204e-4} & \multicolumn{1}{l|}{5.585e-1$\pm$2.578e-2} & \multicolumn{1}{l|}{5.363e-1$\pm$3.698e-2} & \multicolumn{1}{l|}{7.496e-2$\pm$5.147e-3} & \multicolumn{1}{l|}{1.299e-3$\pm$6.210e-4}   \\
\multicolumn{1}{l|}{Steady NS} & \multicolumn{1}{l|}{5.264e-1$\pm$1.013e-2} & \multicolumn{1}{l|}{7.143e-1$\pm$1.325e-2} & \multicolumn{1}{l|}{6.332e-1$\pm$2.345e-2} & \multicolumn{1}{l|}{6.939e-1$\pm$1.064e-3} & \multicolumn{1}{l|}{3.769e-1$\pm$2.367e-2} \\
\midrule
\end{tabular}
\end{adjustbox}
\centering
\scriptsize
\begin{tabular}{@{}ccccc@{}}
\toprule
          & CT                     & CG                       & \textbf{RBF-INT}                            & \textbf{RBF-POL}   \\ \midrule
\multicolumn{1}{l|}{Wave}     & \multicolumn{1}{l|}{1.114e0$\pm$3.214e-2}  & \multicolumn{1}{l|}{1.036e0$\pm$1.054e-2}  & \multicolumn{1}{l|}{2.814e-2$\pm$3.647e-3} & 2.361e-2$\pm$1.598e-2  \\
\multicolumn{1}{l|}{Diffusion}& \multicolumn{1}{l|}{1.860e0$\pm$2.312e-2}  & \multicolumn{1}{l|}{2.721e-2$\pm$1.027e-1} & \multicolumn{1}{l|}{3.066e-4$\pm$9.517e-6} & \colorbox{blue!25}{3.498e-5$\pm$6.547e-6} \\
\multicolumn{1}{l|}{Heat}     & \multicolumn{1}{l|}{4.524e-1$\pm$6.514e-2} & \multicolumn{1}{l|}{2.626e-1$\pm$2.367e-2}  & \multicolumn{1}{l|}{1.157e-3$\pm$1.020e-4} & \colorbox{blue!25}{4.098e-4$\pm$9.621e-6} \\
\multicolumn{1}{l|}{Poisson}   & \multicolumn{1}{l|}{6.348e-1$\pm$3.049e-1} & \multicolumn{1}{l|}{2.334e-1$\pm$5.471e-2} & \multicolumn{1}{l|}{\colorbox{blue!25}{5.259e-4$\pm$6.243e-5}} & 8.942e-4$\pm$ 6.514e-5\\
\multicolumn{1}{l|}{Burgers'}   & \multicolumn{1}{l|}{9.935e-1$\pm$4.512e-2} & \multicolumn{1}{l|}{7.521e-1$\pm$3.249e-2} & \multicolumn{1}{l|}{2.945e-3$\pm$2.354e-4} & \colorbox{blue!25}{3.159e-4$\pm$2.146e-5} \\
\multicolumn{1}{l|}{Steady NS}  & \multicolumn{1}{l|}{5.460e-1$\pm$2.357e-2}  & \multicolumn{1}{l|}{4.867e-1$\pm$3.654e-2} & \multicolumn{1}{l|}{2.991e-1$\pm$6.514e-2} & \colorbox{blue!25}{2.567e-1$\pm$6.217e-2} \\
\midrule
\end{tabular}

\end{table}

\subsection{Complete results for Table~\ref{tab: inverse}}
\begin{table}[h]
\caption{Full Benchmark results on the Inverse problems in $\ell^2$ error. * indicates problems with noises added to the data.}
\label{tab: full inverse}
\centering
\tiny
\begin{adjustbox}{width=\columnwidth,center}
\begin{tabular}{lllll}
\toprule
                                                          & FF                             & SF                            & \textbf{RBF-INT}                       & \textbf{RBF-POL}  \\ \midrule
\multicolumn{1}{l|}{I-Burgers'}  & \multicolumn{1}{l|}{2.391e-2$\pm$9.647e-4} & \multicolumn{1}{l|}{2.436e-2$\pm$4.678e-3} & \multicolumn{1}{l|}{1.741e-2$\pm$6.571e-3} & \colorbox{blue!25}{1.575e-2$\pm$9.369e-4} \\
\multicolumn{1}{l|}{I-Lorenz}   & \multicolumn{1}{l|}{6.516e-3$\pm$7.651e-4}  & \multicolumn{1}{l|}{6.390e-3$\pm$6.214e-4} & \multicolumn{1}{l|}{6.080e-3$\pm$3.697e-4} & \colorbox{blue!25}{5.991e-3$\pm$2.312e-4} \\ 
\multicolumn{1}{l|}{I-Burgers'*}  & \multicolumn{1}{l|}{2.509e-2$\pm$6.324e-3} & \multicolumn{1}{l|}{2.913e-2$\pm$2.698e-3} & \multicolumn{1}{l|}{1.993e-2$\pm$3.621e-3} & \colorbox{blue!25}{1.753e-2$\pm$5.632e-3} \\
\multicolumn{1}{l|}{I-Lorenz*}   & \multicolumn{1}{l|}{7.934e-3$\pm$8.651e-4}  & \multicolumn{1}{l|}{6.856e-3$\pm$6.363e-4} & \multicolumn{1}{l|}{6.699e-3$\pm$5.201e-4} & \colorbox{blue!25}{6.342e-3$\pm$8.614e-4} \\ 
\midrule
\end{tabular}
\end{adjustbox}
\end{table}

\newpage

\section{Ablation Study}
\label{appendix: ablation study}
In this section, we show some additional experiments on our RBF feature mapping including investigations on the Number of RBFs, Number of Polynomials and different RBF types.

\subsection{Number of RBFs}
\begin{figure}[h]
\begin{center}
\centerline{\includegraphics[height=6cm]{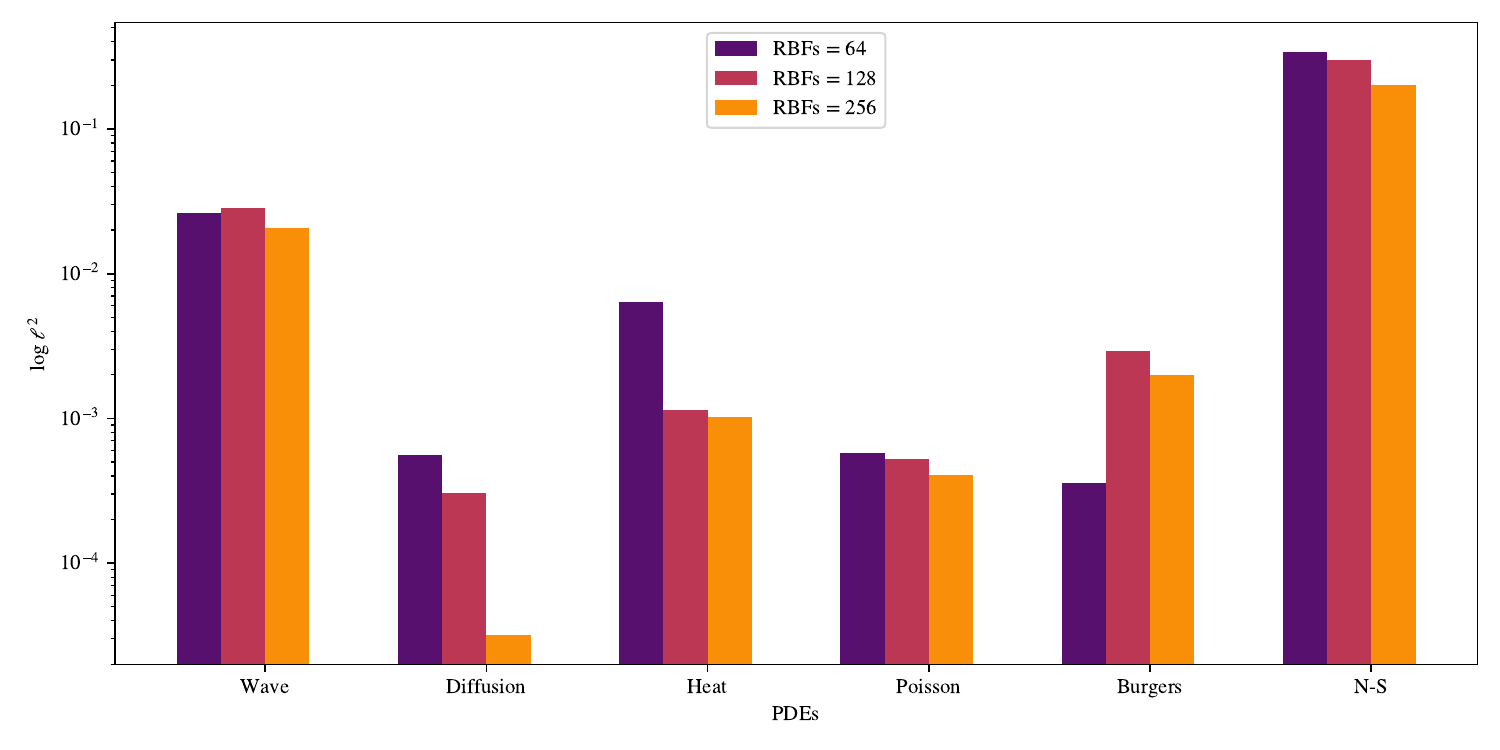}}
\caption{Ablation study on different number of RBFs}
\label{ablation: rbf number}
\end{center}
\end{figure}

Figure~\ref{ablation: rbf number} has shown generally more RBFs (256) yield better results. It however does demand a higher memory and can be slow in some cases.  It shows in the Diffusion equation, with 256 RBFs, the error reduces quite significantly. Otherwise, it only has limited improvements because the error is already very low. We use 128 RBFs in general case for a better performance-speed tradeoff.

\subsection{Number of Polynomials}
\begin{figure}[h]
\begin{center}
\centerline{\includegraphics[height=6cm]{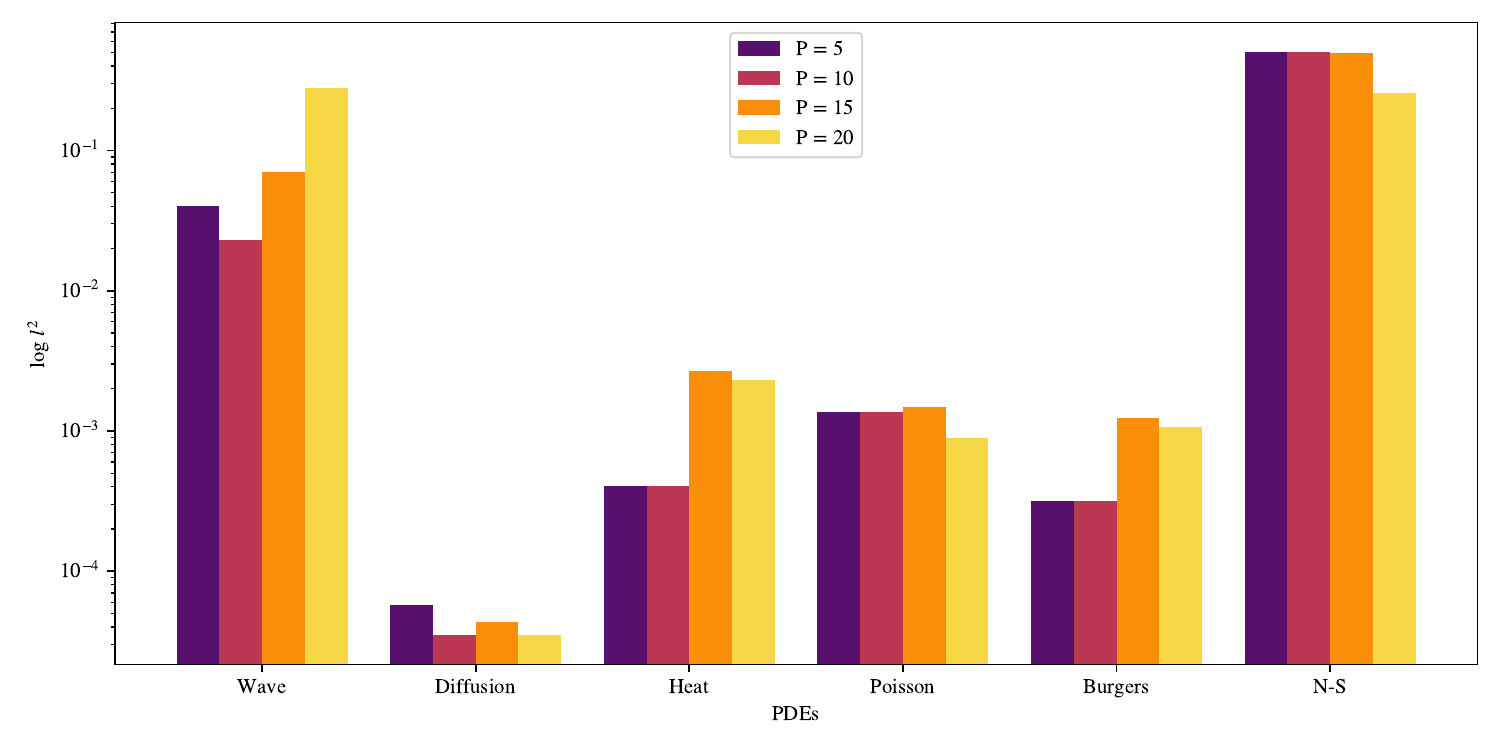}}
\caption{Ablation study on different number of polynomials}
\label{Fig: abla-pol}
\end{center}
\end{figure}

Figure~\ref{Fig: abla-pol} shows an ablation study of how the number of polynomials in feature mappings influences performance in PDEs. It has shown RBF feature mapping with 20 polynomials has achieved best results in the Diffusion equation, Poisson equation and N-S equation. And 10 polynomial terms are better in Heat equation and Burgers' equation, thought its performance is matching with only 5 polynomials. 

\subsection{Different Types of RBFs}
Following Table~\ref{tab: RBF types} are common positive definite Radial Basis Functions.
\begin{table}[ht]
\caption{Types of Radial Basis function and their formulation. $\mathbf{x}-\mathbf{c}$ is shorten as r.}
\centering
\label{tab: RBF types}
\begin{tabular}{|l|l|}
\hline
Type                   & Radial function \\ \hline
Cubic                  & $r^3 $              \\ \hline
TPS(Thin Plate Spline) & $r^2log(r)$               \\ \hline
GA(Gaussian)           & $e^{-r^2/\sigma^2}$               \\ \hline
MQ(Multiquadric)       & $\sqrt{1+r^2} $              \\ \hline
IMQ(Inverse MQ)        & $1/\sqrt{1+r^2} $               \\ \hline
\end{tabular}
\end{table}

\begin{figure}[h]
\begin{center}
\centerline{\includegraphics[height=6cm]{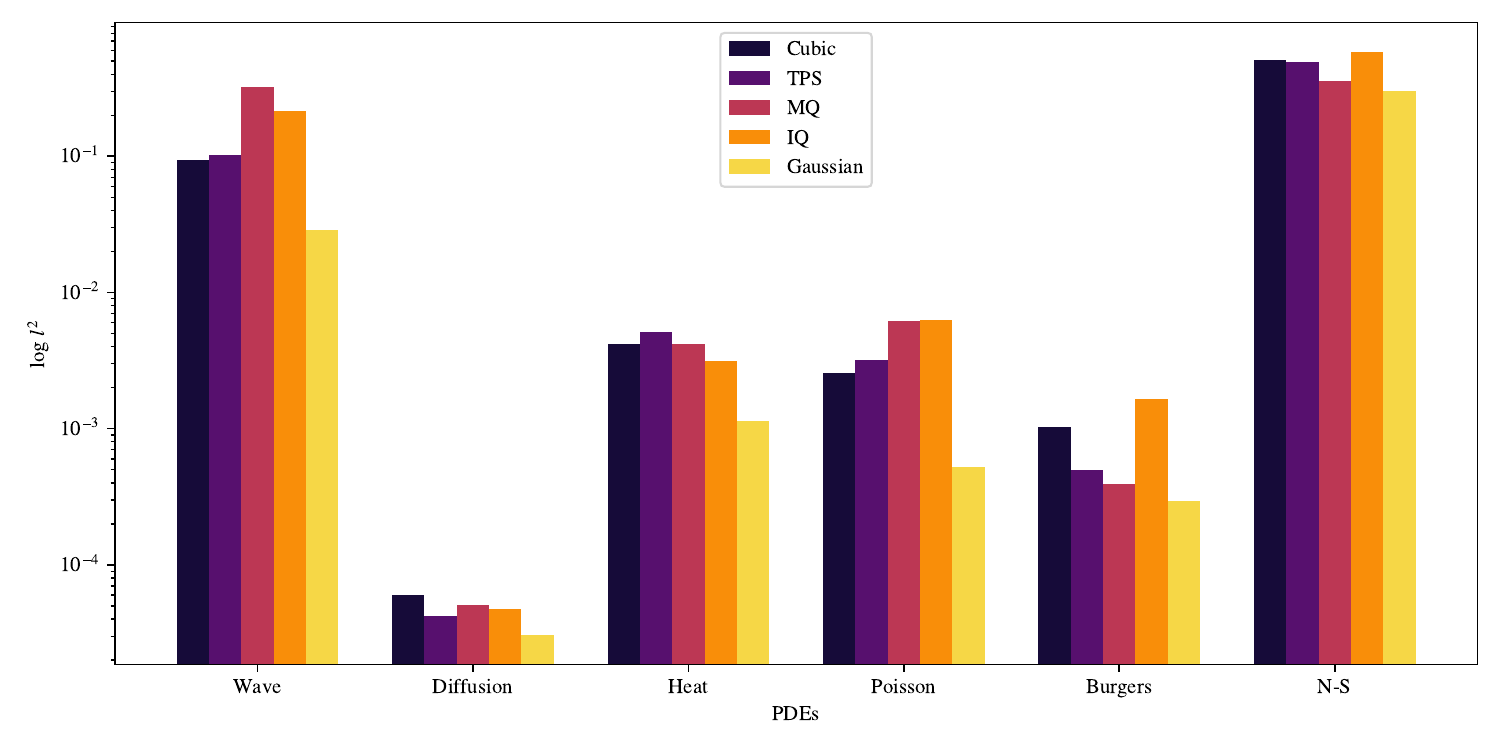}}
\caption{Ablation study on different types of RBFs}
\label{Fig: RBF types}
\end{center}
\end{figure}

The Figure~\ref{Fig: RBF types} has shown Gaussian RBF is dominating all types of PDEs. However other types of RBF are in similar performance. We generally prefer Gaussian RBF in all cases due to its nice properties.

\section{Convergence, Complexity and Scalability analysis}
\label{appendix: complexity}


\begin{figure}[h!]
\centering
\begin{subfigure}{0.45\textwidth}
    \includegraphics[width=\textwidth]{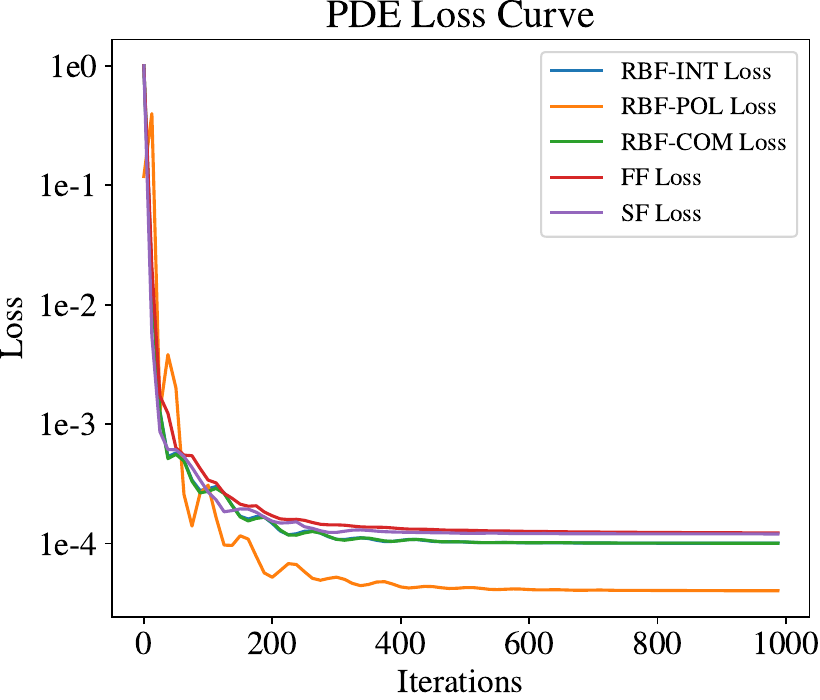}
    \caption{}
\vspace{-0.1in}
\end{subfigure} 
\hspace{0.4in}
\begin{subfigure}{0.45\textwidth}
    \includegraphics[width=\textwidth]{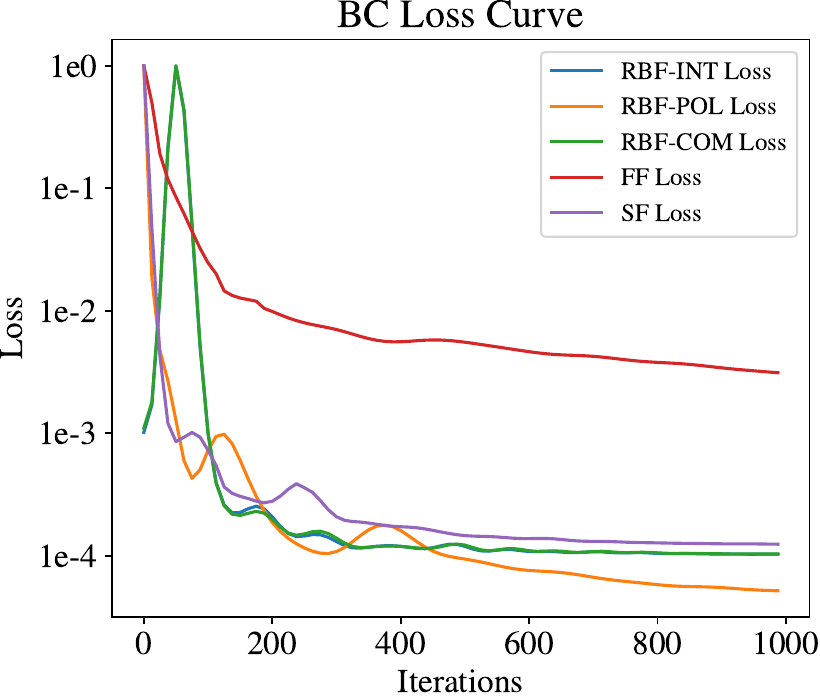}
    \caption{}
\vspace{-0.1in}
\end{subfigure}
\caption{Loss curves of PDE and BC loss on Diffusion equation.}
\vspace{-0.2in}
\label{figure: convergence}
\end{figure}

Convergence analysis on the Diffusion equation is shown in Figure~\ref{figure: convergence}. Our methods not only show better convergences overall, but also show a better adjustment with the boundary conditions.

Although all feature mapping methods are similar in computational complexity, for completeness, we include the detailed computational complexity of the feature layers that map 128 features and 4 fully connected layers with 50 neurons each, in Table~\ref{tab: complexity}. 

\begin{table*}[ht]
\caption{Computational complexity}
\center
\begin{adjustbox}{width=\columnwidth,center}
\begin{tabular}{|l|l|l|l|l|l|l|l|}
\hline
       & FF     & SF     & RBF-INT & RBF-POL-5 & RBF-POL-10 & RBF-POL-15 & RBF-POL-20 \\ \hline
FLOPs  & 139.5M & 142.1M & 139.5M  & 142.5M    & 145.0M     & 147.5M     & 150.0M     \\ \hline
Params & 14.2k  & 14.3k  & 14.2k   & 14.5k     & 14.7k      & 14.9k      & 15.2k      \\ \hline
\end{tabular}
\end{adjustbox}
\label{tab: complexity}
\end{table*}

\begin{figure}[ht]
\begin{center}
\centerline{\includegraphics[height=6cm]{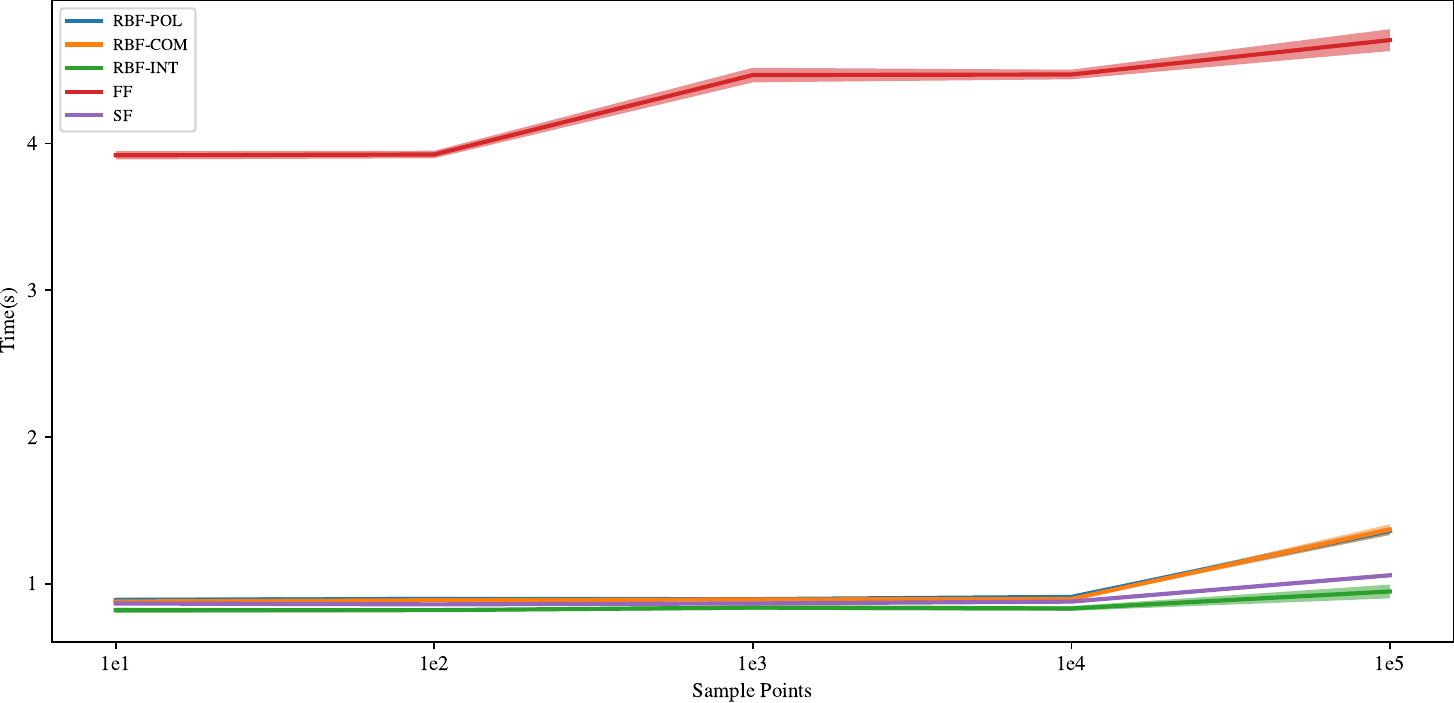}}
\caption{Time consumption on different numbers of sample points with different feature mapping methods}
\label{Fig: time}
\end{center}
\end{figure}

Due to software optimisation and package compatibility, the feature mapping methods can have very different computational efficiency in training.
To demonstrate, we run the above models on different numbers of sample points on Diffusion equation for 3 times in different random seeds for 1 epoch. RBF-COM stands for compact support RBF, the support distance $\xi = 4$ for all cases, and RBF-POL uses 20 polynomials in Figure~\ref{Fig: time}.\\
The time consumed by Fourier Features is noticeably higher than other methods. All methods have similar runtime for sample points less than $1e4$, that is because all sample points computed are within one single GPU parallelisation capacity.

\newpage

\section{Benchmark PDEs and Boundary conditions}
\label{Append: Benchmark}

\subsection{Wave Equation}
The one-dimensional Wave Equation is given by:
\begin{equation}
u_{tt} - 4u_{xx}=0,
\end{equation}
In the domain of: 
\begin{equation}
(x, t) \in \Omega \times T = [0,1] \times[0,1],
\end{equation}
Boundary condition:
\begin{equation}
u(0, t) = u(1, t) = 0,
\end{equation}
Initial condition:
\begin{align}
u(x, 0) &= \sin(\pi x) + \frac{1}{2}\sin(4\pi x), \\
u_t &= 0,  \\
\end{align}
The analytical solution of the equation is:
\begin{equation}
u(x, t) = \sin(\pi x)cos(2\pi t) + \frac{1}{2}\sin(4\pi x)cos(8\pi t),
\end{equation}

\subsection{Diffusion Equation}
The one-dimensional Diffusion Equation is given by:
\begin{equation}
u_t - u_{xx} + e^{-t}(sin(\pi x) + \pi^2sin(\pi x))= 0,
\end{equation}
In the domain of: 
\begin{equation}
(x, t) \in \Omega \times T = [-1,1] \times[0,1],
\end{equation}
Boundary condition:
\begin{equation}
u(-1, t) = u(1, t) = 0,
\end{equation}
Initial condition:
\begin{align}
u(x, 0) = sin(\pi x),
\end{align}
The analytical solution of the equation is:
\begin{equation}
u(x, t) = e^tsin(\pi x),
\end{equation}
where $\alpha = 0.4, L = 1, n = 1$

\subsection{Heat Equation}
\label{equation: heat}
The two-dimensional Heat Equation is given by:
\begin{equation}
\quad u_{t}-\frac{1}{(500 \pi)^2} u_{x x}-\frac{1}{\pi^2} u_{y y}=0,
\end{equation}

In the domain of: 
\begin{equation}
(\mathbf{x}, t) \in \Omega \times T = [0,1]^2 \times[0,5],
\end{equation}

Boundary condition:
\begin{equation}
u(x, y, t)=0,
\end{equation}

Initial condition:
\begin{align}
u(x, y, 0)=\sin (20 \pi x) \sin (\pi y),
\end{align}

\subsection{Poisson Equation}
The two-dimensional Poisson Equation is given by:
\begin{equation}
-\Delta u = 0,
\end{equation}
In the domain of: 
\begin{equation}
\mathbf{x} \in \Omega = \Omega_{rec} \backslash R_{i},
\end{equation}
where
\begin{align}
\Omega_{rec} &= [-0.5, 0.5]^2, \\
R_1& =[(x, y):(x-0.3)^2+(y-0.3)^2 \leq 0.1^2], \\
R_2& =[(x, y):(x+0.3)^2+(y-0.3)^2 \leq 0.1^2], \\
R_3& =[(x, y):(x-0.3)^2+(y+0.3)^2 \leq 0.1^2], \\
R_4& =[(x, y):(x+0.3)^2+(y+0.3)^2 \leq 0.1^2].
\end{align}
Boundary condition:
\begin{align}
u &= 0, x \in \partial R_i, \\
u &= 1, x \in \partial \Omega_{rec},
\end{align}

\subsection{Burgers Equation}
The one-dimensional Burgers' Equation is given by:
\begin{equation}
u_t+u u_x=\nu u_{x x},
\end{equation}
In the domain of: 
\begin{equation}
(x, t) \in \Omega=[-1,1] \times[0,1],
\end{equation}
Boundary condition:
\begin{equation}
u(-1, t)=u(1, t)=0,
\end{equation}
Initial condition:
\begin{align}
u(x, 0)=-\sin \pi x,
\end{align}
where $\nu=\frac{0.01}{\pi}$

\subsection{Steady NS}
The steady incompressible Navier Stokes Equation is given by:
\begin{align}
\nabla \cdot \mathbf{u} & =0 ,\\
\mathbf{u} \cdot \nabla \mathbf{u}+\nabla p-\frac{1}{\operatorname{Re}} \Delta \mathbf{u} & =0, \\
\end{align}

In the domain(back step flow) of: 
\begin{equation}
\mathbf{x} \in \Omega=[0,4] \times[0,2] \backslash\left([0,2] \times[1,2] \cup R_i\right),
\end{equation}
Boundary condition:
\begin{align}
\text{no-slip condition:} \quad \mathbf{u} &= 0,\\
\text{inlet:} \quad u_x &= 4y(1-y), u_y = 0,\\
\text{outlet:} \quad p &= 0,
\end{align}
where $Re = 100$

\subsection{nD Poisson Equation}
The nth-dimensional Poisson Equation is given by:
\begin{equation}
-\Delta u=\frac{\pi^2}{4} \sum_{i=1}^n \sin \left(\frac{\pi}{2} x_i\right),
\end{equation}
In the domain of: 
\begin{equation}
x\in \Omega =[0,1]^n,
\end{equation}
Boundary condition:
\begin{equation}
u = 0,
\end{equation}
The analytical solution of the equation is:
\begin{equation}
u=\sum_{i=1}^n \sin \left(\frac{\pi}{2} x_i\right),
\end{equation}

\subsection{Inverse Burgers' Equation}
The one-dimensional Inverse Burgers' Equation is given by:
\begin{equation}
u_t+\mu_1 u u_x=\mu_2 u_{x x},
\end{equation}
In the domain of: 
\begin{equation}
(x, t) \in \Omega=[-1,1] \times[0,1],
\end{equation}
Boundary condition:
\begin{equation}
u(-1, t)=u(1, t)=0,
\end{equation}
Initial condition:
\begin{align}
u(x, 0)=-\sin \pi x,
\end{align}
where $\mu_1 = 1$ and $\mu_2 =\frac{0.01}{\pi}$

\subsection{Inverse Lorenz Equation}
\label{equation: lorenz}
The 1st-order three-dimensional Lorenz Equation is given by:
\begin{equation}
\begin{aligned}
\frac{dx}{dt} & = \alpha(y - x) ,\\
\frac{dy}{dt} & = x(\rho - z) - y ,\\
\frac{dz}{dt} & = xy - \beta z ,
\end{aligned}
\end{equation}

where $\alpha = 10$, $\beta = \frac{8}{3}$, $\rho = 15$ and the initial points are $x_0 = 0$, $y_0 = 1$, $z_0=1.05$.

\newpage
\section{Visualisations of PDEs solution}
\label{Append: Qualitative}

\begin{figure}[h]
\begin{center}
\begin{tabular}{c} \hspace*{-0.4cm}
\includegraphics[height=19cm]{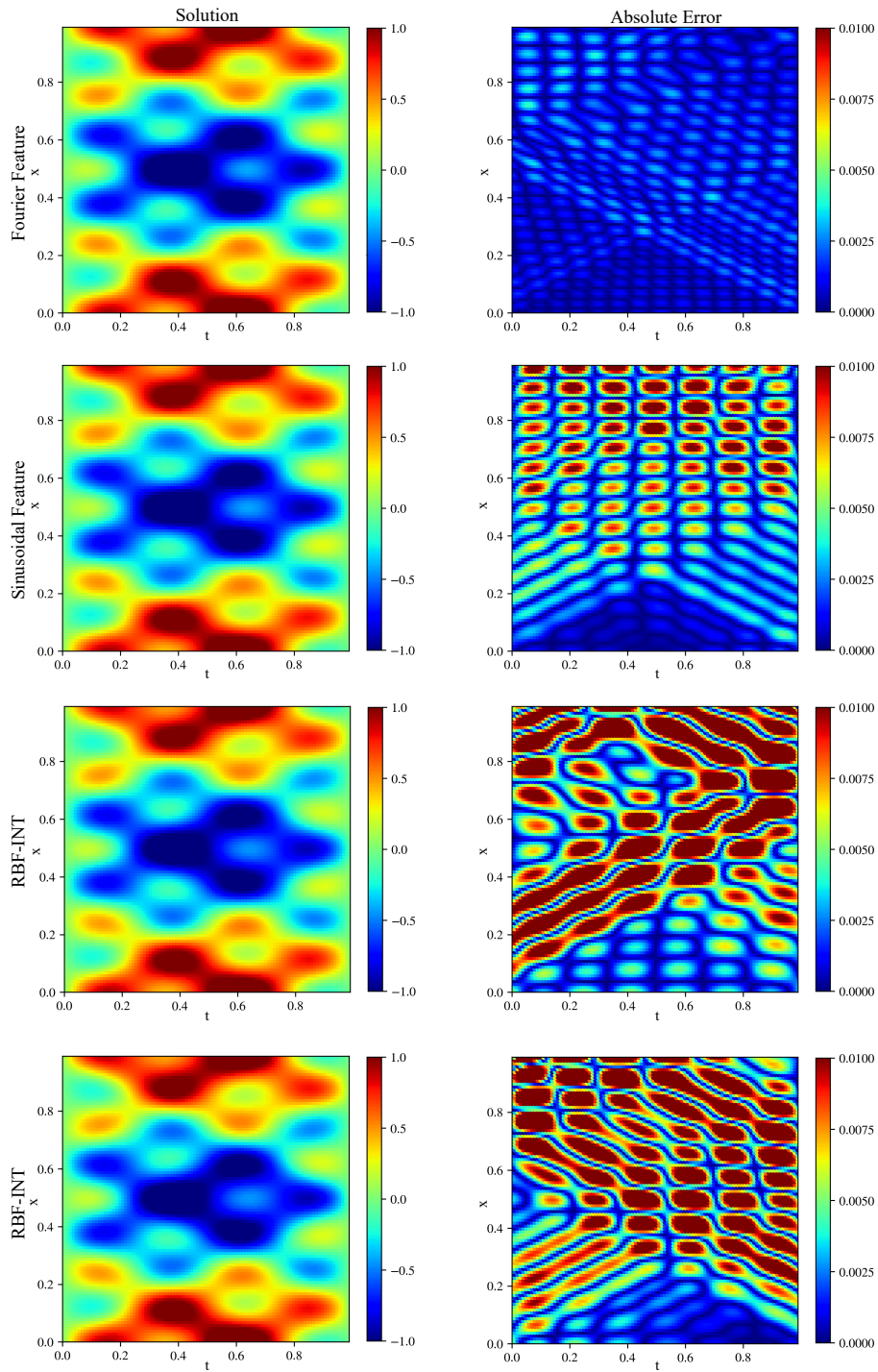}
\end{tabular}
\end{center}
\caption[example] 
{\label{fig:wave qualitative} 
Wave equation}
\end{figure} 

\newpage

\begin{figure}[h]
\begin{center}
\begin{tabular}{c} \vspace{-30pt}
\includegraphics[height=15cm]{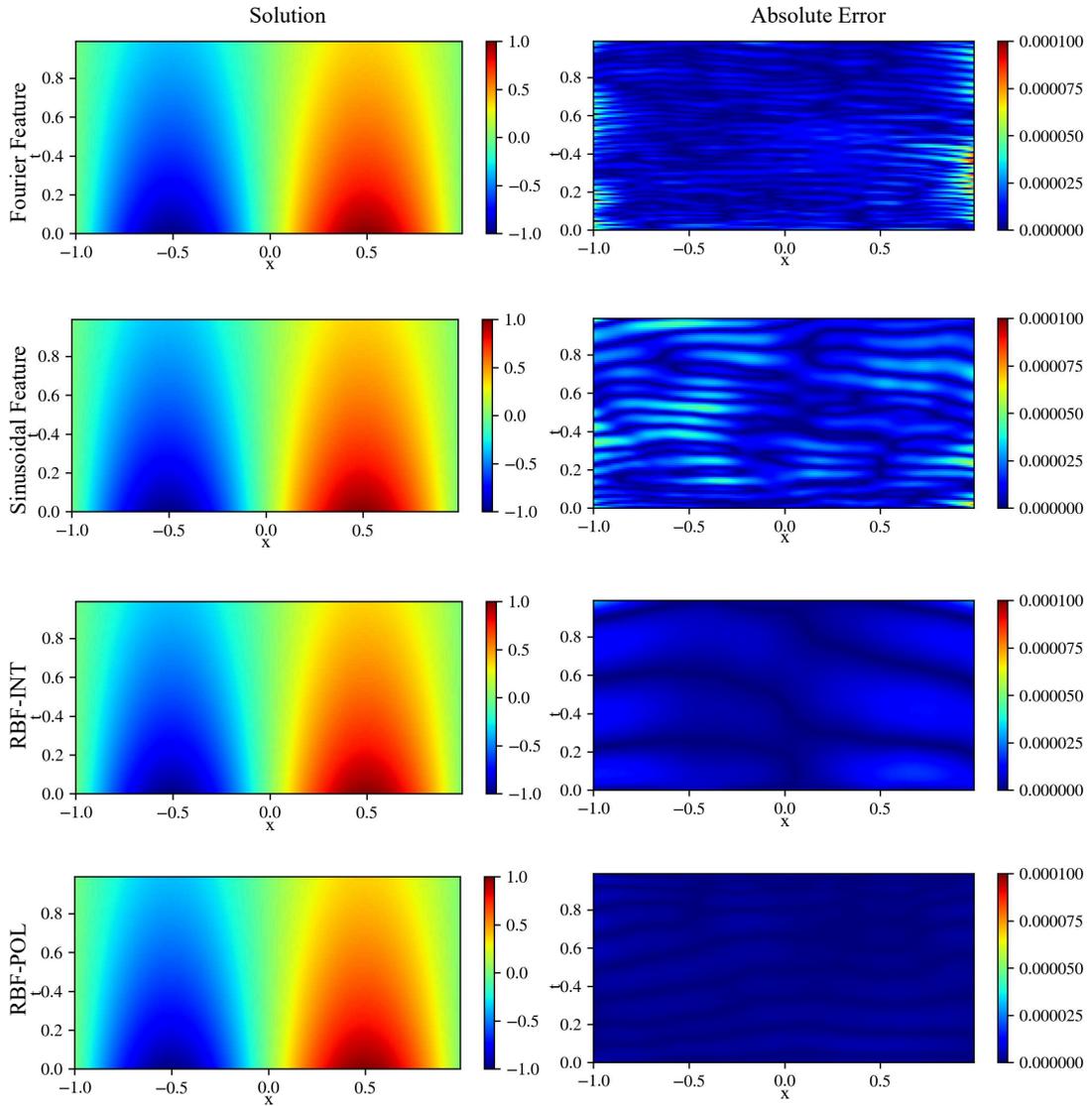}
\end{tabular}
\end{center}
\vspace{50pt}
\caption[example] 
{\label{fig:diffusion qualitative} 
Diffusion equation}
\end{figure} 

\newpage

\begin{figure}[h]
\begin{center}
\begin{tabular}{c} \vspace{-60pt}\hspace*{-2.5cm}
\includegraphics[height=18cm]{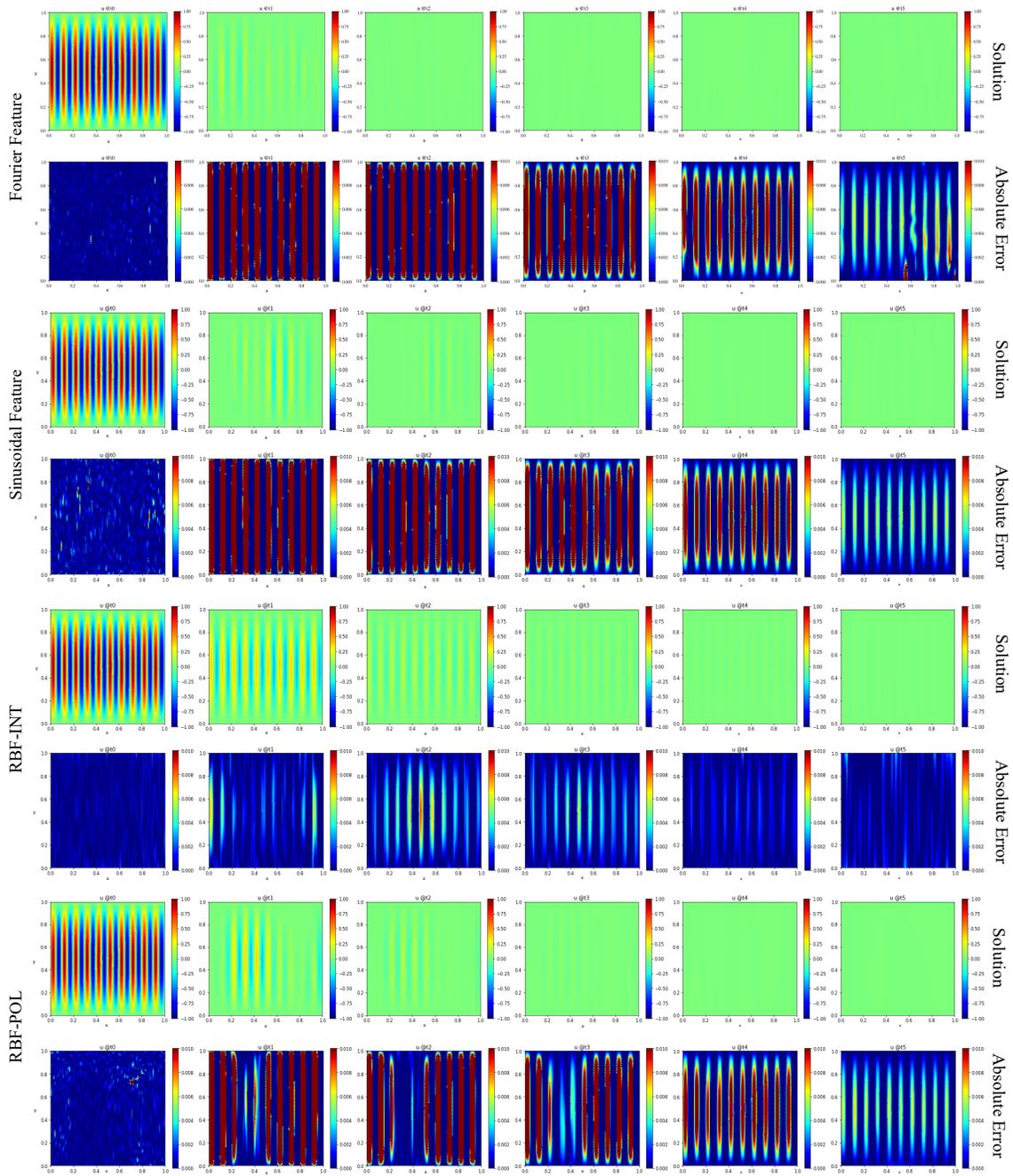}
\end{tabular}
\end{center}
\vspace{50pt}
\caption[example] 
{\label{fig:heat qualitative} 
Heat equation}
\end{figure}

\newpage

\begin{figure}[h]
\begin{center}
\begin{tabular}{c} \vspace{-40pt}\hspace*{-1.5cm}
\includegraphics[height=18cm]{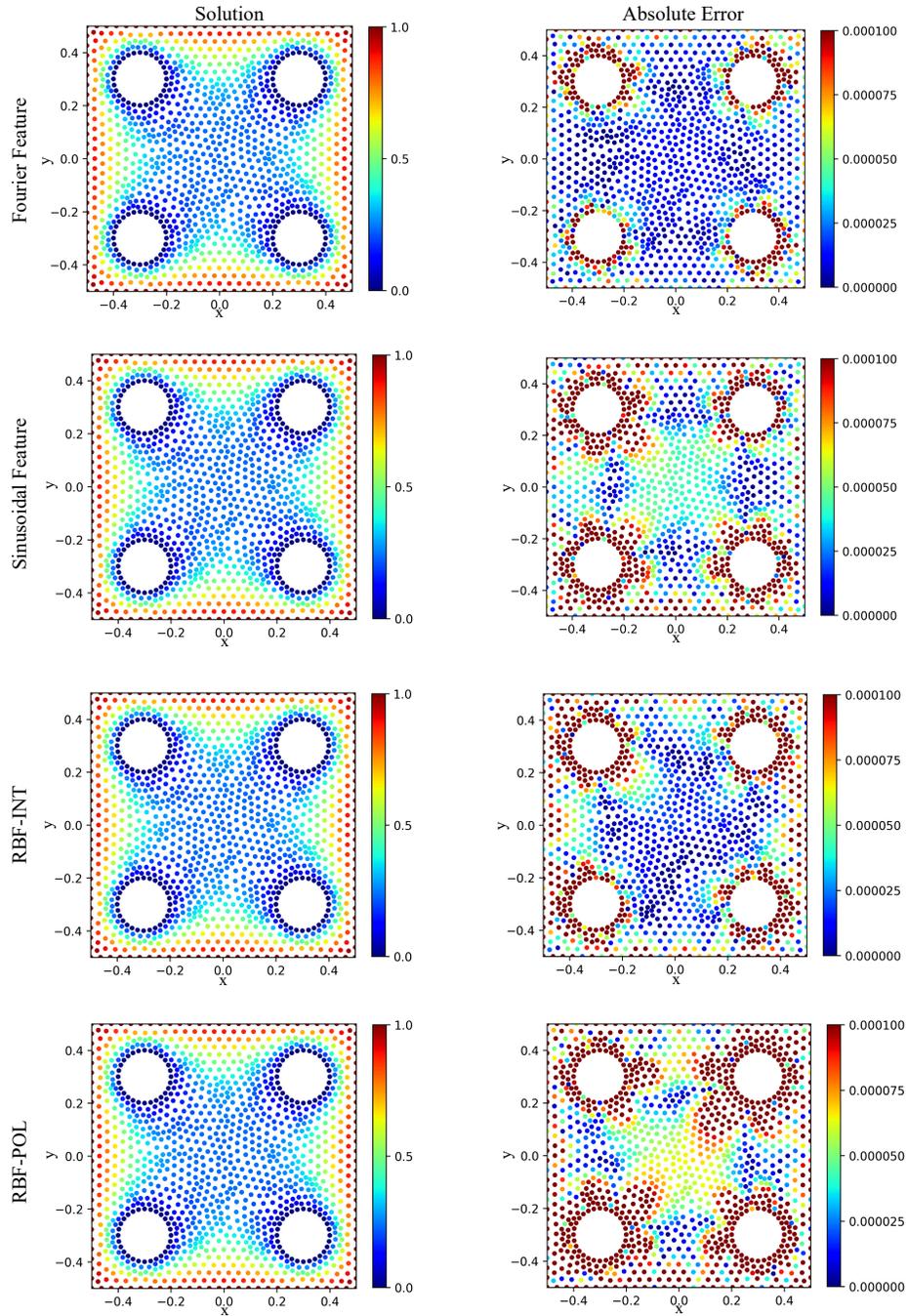}
\end{tabular}
\end{center}
\vspace{30pt}
\caption[example] 
{\label{fig:poisson qualitative} 
Poisson equation}
\end{figure}

\newpage

\begin{figure}[h]
\begin{center}
\begin{tabular}{c} \vspace{-40pt}\hspace*{-1.5cm}
\includegraphics[height=15cm]{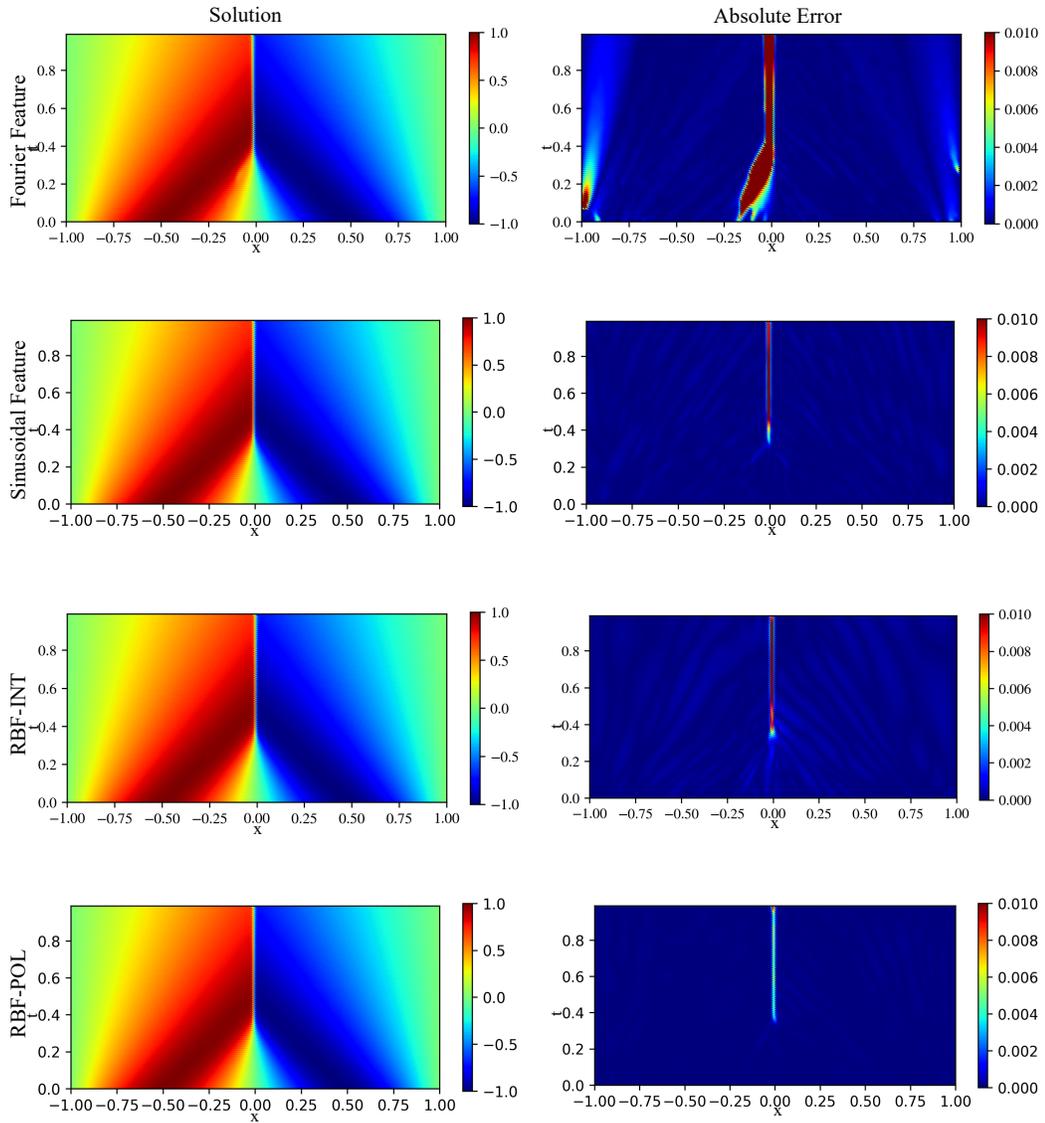}
\end{tabular}
\end{center}
\vspace{30pt}
\caption[example] 
{\label{fig:burgers qualitative} 
 Burgers' equation}
\end{figure} 

\newpage

\begin{figure}[h]
\begin{center}
\begin{tabular}{c} \vspace{-30pt}\hspace*{-1.5cm}
\includegraphics[height=18cm]{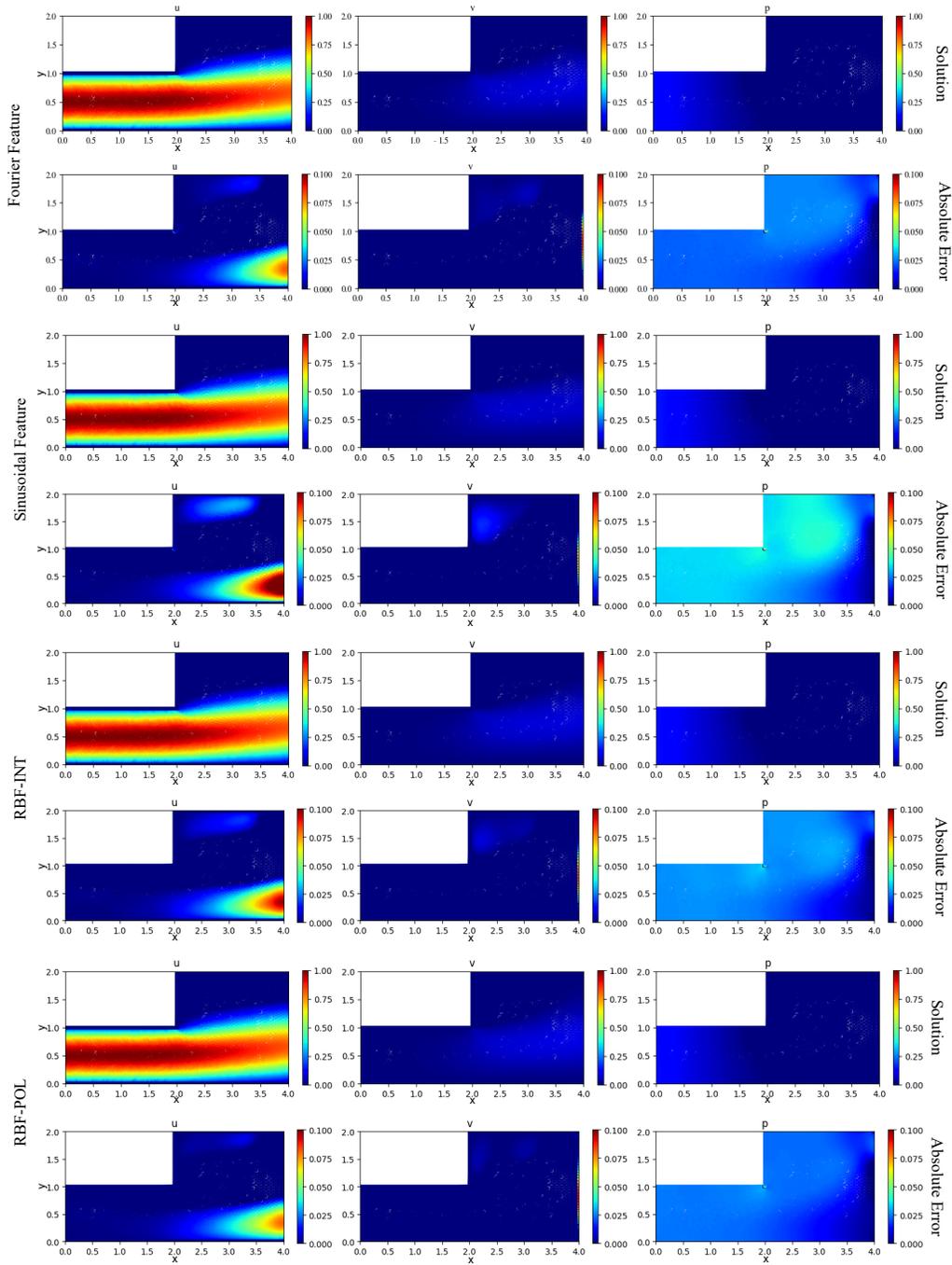}
\end{tabular}
\end{center}
\vspace{30pt}
\caption[example] 
{\label{fig:ns_equation qualitative} 
 Navier-Stokes equation}
\end{figure} 

\newpage

\begin{figure}[h]
\begin{center}
\begin{tabular}{c} \vspace{-10pt}\hspace*{0.5cm}
\includegraphics[height=16cm]{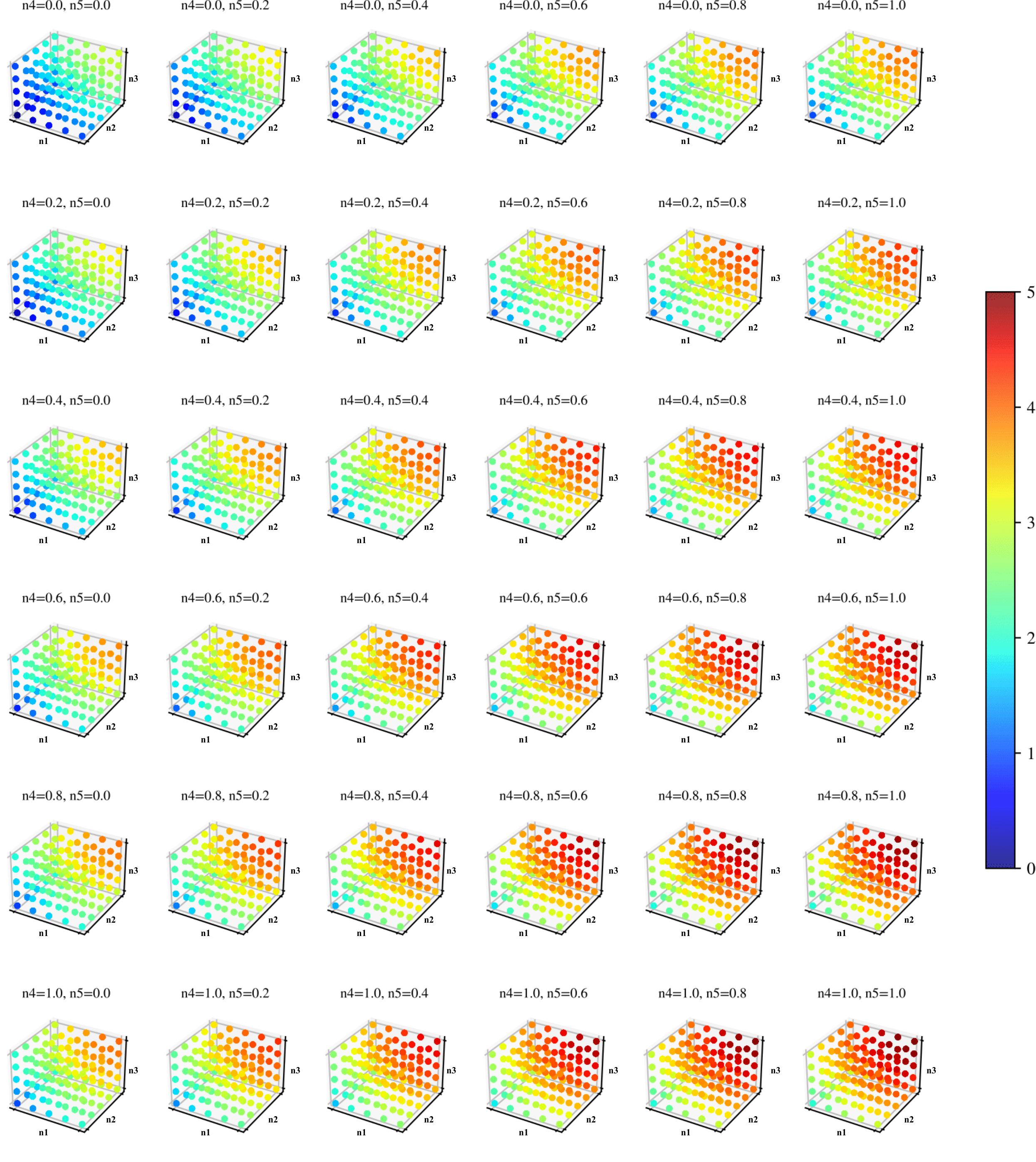}
\end{tabular}
\end{center}
\vspace{20pt}
\caption[example] 
{\label{fig:ndpoisson_gt qualitative} 
 nD Poisson equation Ground Truth, when n = 5, the x,y,z direction of the cube is the 1st, 2nd and 3rd dimension, respectively, the rows direction of the image is the 4th dimension and the column direction of the image is the 5th dimension.}
\end{figure} 

\newpage

\begin{figure}[h]
\begin{center}
\begin{tabular}{c} \vspace{-10pt}\hspace*{0.5cm}
\includegraphics[height=16cm]{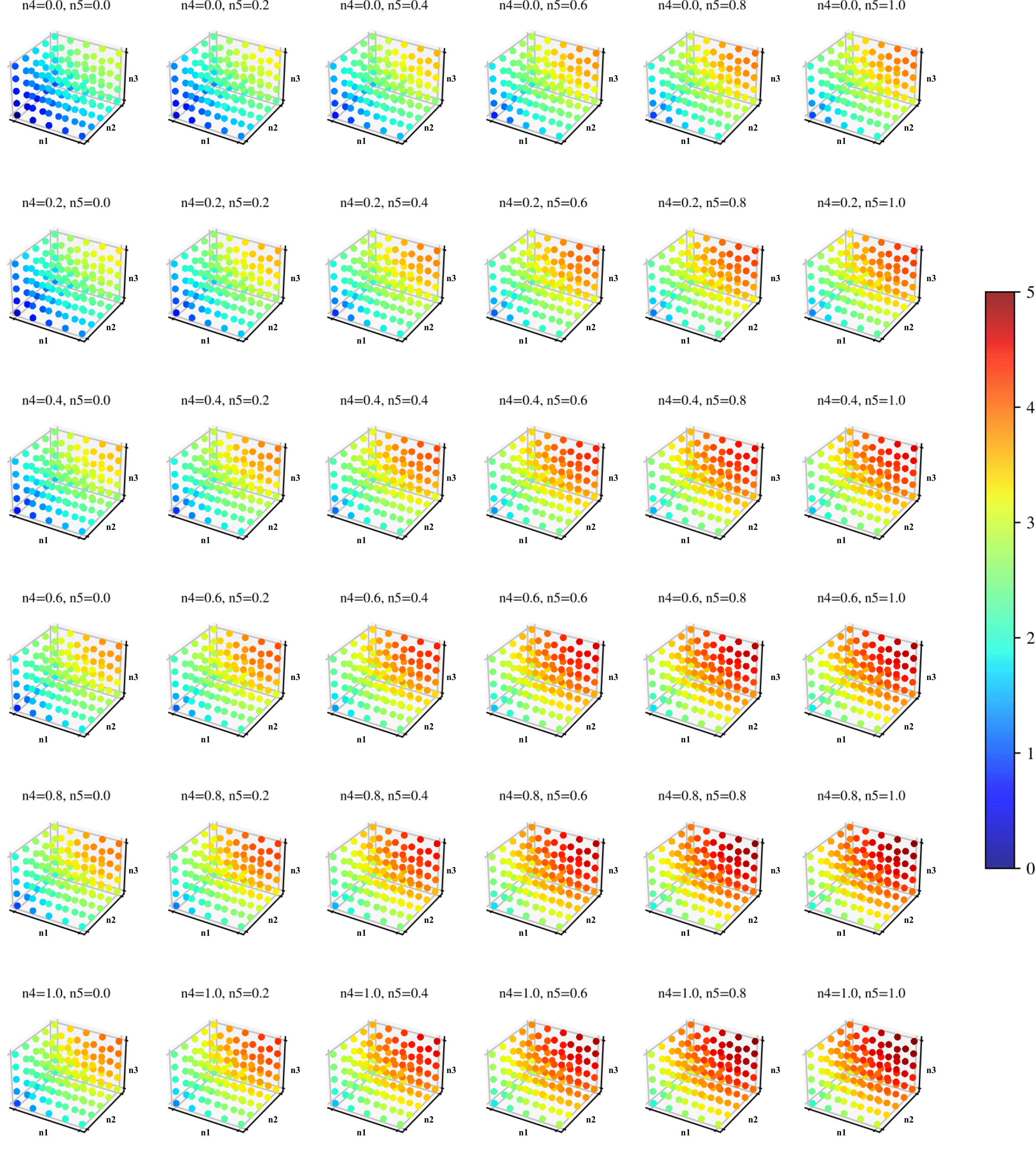}
\end{tabular}
\end{center}
\vspace{20pt}
\caption[example] 
{\label{fig:ndpoisson_gt qualitative} 
 5D Poisson Equation solved by RBF feature mapping.}
\end{figure} 
\newpage
\begin{figure}[h]
\begin{center}
\begin{tabular}{c} \vspace{-10pt}\hspace*{0.5cm}
\includegraphics[height=16cm]{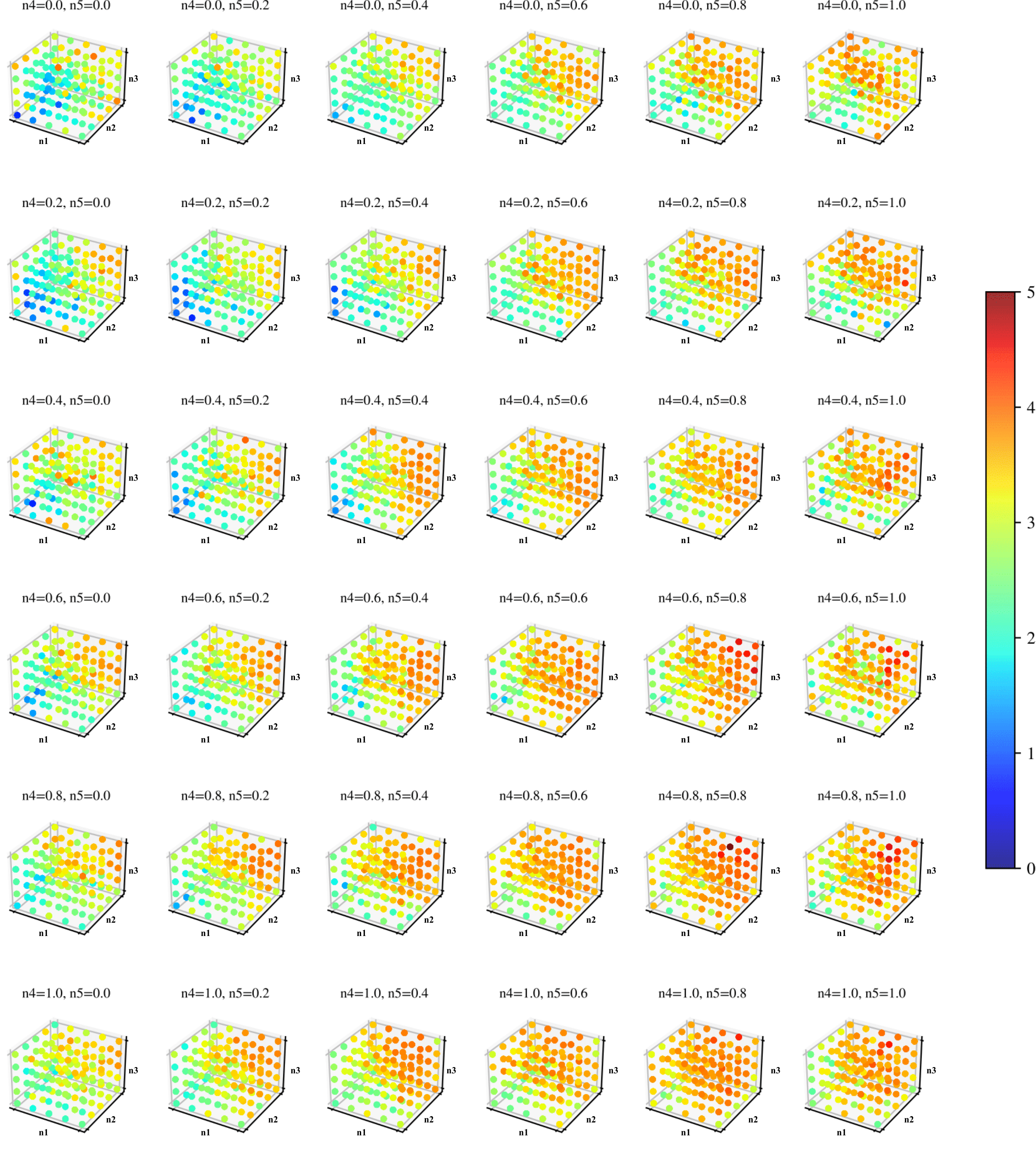}
\end{tabular}
\end{center}
\vspace{20pt}
\caption[example] 
{\label{fig:ndpoisson_ff qualitative} 
 5D Poisson Equation solved by Fourier feature mapping.}
\end{figure} 

\newpage

\begin{figure}[h]
\begin{center}
\begin{tabular}{c}
\includegraphics[height=18cm]{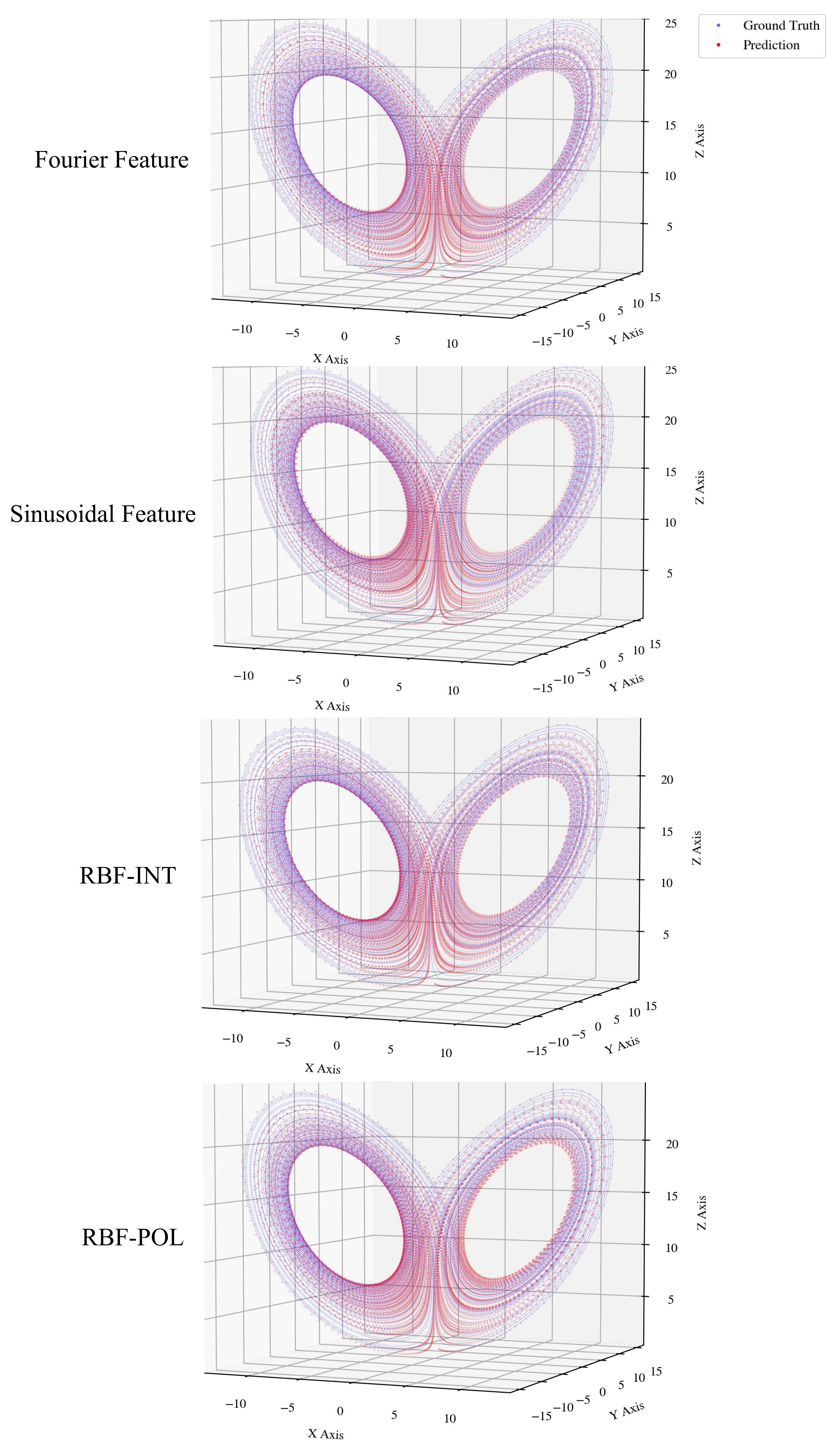}
\end{tabular}
\end{center}
\caption[example] 
{\label{fig:lorenz} 
 Visualisation of the Lorenz system with coefficients predicted by different feature mapping. A slight change of any of the coefficients will result visible deviations. (Zoom in for better trajectory visualisation.)}
\end{figure}

\newpage

\newpage

\section*{NeurIPS Paper Checklist}

\begin{enumerate}

\item {\bf Claims}
    \item[] Question: Do the main claims made in the abstract and introduction accurately reflect the paper's contributions and scope?
    \item[] Answer: \answerYes{} 
    \item[] Justification: Our paper is based on two scopes, one is the properties revealed by our theoretical work on the training dynamics of the PINNs with feature mapping, the other one is the limitation of Fourier Features. We reflect this in both abstract and introduction. Particularly, we include a list of contributions to clarify our work.

\item {\bf Limitations}
    \item[] Question: Does the paper discuss the limitations of the work performed by the authors?
    \item[] Answer: \answerYes{} 
    \item[] Justification: A detailed limitation of our work is included in Section~\ref{section: limitation}

\item {\bf Theory Assumptions and Proofs}
    \item[] Question: For each theoretical result, does the paper provide the full set of assumptions and a complete (and correct) proof?
    \item[] Answer: \answerYes{} 
    \item[] Justification: The assumptions for Theorem 3.1~\ref{Prop: ck} and Theorem 3.2~\ref{theorem: ntk} are made at the beginning of Page 5. The full proofs are followed in Appendices~\ref{Append: inject},~\ref{Append: ck} and~\ref{Append: ntk}.

    \item {\bf Experimental Result Reproducibility}
    \item[] Question: Does the paper fully disclose all the information needed to reproduce the main experimental results of the paper to the extent that it affects the main claims and/or conclusions of the paper (regardless of whether the code and data are provided or not)?
    \item[] Answer: \answerYes{} 
    \item[] Justification: In addition to setup details in~\ref{section: experiements}, we included inplementation details, software \& hardware used to carry out our experiments in Appendix~\ref{appendix: reproducibility}.

\item {\bf Open access to data and code}
    \item[] Question: Does the paper provide open access to the data and code, with sufficient instructions to faithfully reproduce the main experimental results, as described in supplemental material?
    \item[] Answer: \answerYes{} 
    \item[] Justification: See attached zip file.

\item {\bf Experimental Setting/Details}
    \item[] Question: Does the paper specify all the training and test details (e.g., data splits, hyperparameters, how they were chosen, type of optimizer, etc.) necessary to understand the results?
    \item[] Answer: \answerYes{} 
    \item[] Justification: See~\ref{appendix: reproducibility}.

\item {\bf Experiment Statistical Significance}
    \item[] Question: Does the paper report error bars suitably and correctly defined or other appropriate information about the statistical significance of the experiments?
    \item[] Answer: \answerYes{} 
    \item[] Justification: Many key experiments are run multiple times in different seeds. See Figure~\ref{fig:wave qualitative} right, and Appendix~\ref{Append: full results} with full results with mean and variance.

\item {\bf Experiments Compute Resources}
    \item[] Question: For each experiment, does the paper provide sufficient information on the computer resources (type of compute workers, memory, time of execution) needed to reproduce the experiments?
    \item[] Answer: \answerYes{} 
    \item[] Justification: Hardware details are included in Appendix~\ref{appendix: reproducibility}. Time for execution is included in Appendix~\ref{appendix: complexity}.
    
\item {\bf Code Of Ethics}
    \item[] Question: Does the research conducted in the paper conform, in every respect, with the NeurIPS Code of Ethics \url{https://neurips.cc/public/EthicsGuidelines}?
    \item[] Answer: \answerYes{} 
    \item[] Justification: We carefully checked and respected all the Code of Ethics.

\item {\bf Broader Impacts}
    \item[] Question: Does the paper discuss both potential positive societal impacts and negative societal impacts of the work performed?
    \item[] Answer: \answerYes{} 
    \item[] Justification: See Appendix~\ref{appendix: impact}.
    
\item {\bf Safeguards}
    \item[] Question: Does the paper describe safeguards that have been put in place for responsible release of data or models that have a high risk for misuse (e.g., pretrained language models, image generators, or scraped datasets)?
    \item[] Answer: \answerNA{} 
    \item[] Justification: \answerNA{}

\item {\bf Licenses for existing assets}
    \item[] Question: Are the creators or original owners of assets (e.g., code, data, models), used in the paper, properly credited and are the license and terms of use explicitly mentioned and properly respected?
    \item[] Answer:\answerYes{} 
    \item[] Justification: All used third-party code and packages are cited and open-sourced. The license for each package are included when citing.

\item {\bf New Assets}
    \item[] Question: Are new assets introduced in the paper well documented and is the documentation provided alongside the assets?
    \item[] Answer: \answerNA{} 
    \item[] Justification: \answerNA{}

\item {\bf Crowdsourcing and Research with Human Subjects}
    \item[] Question: For crowdsourcing experiments and research with human subjects, does the paper include the full text of instructions given to participants and screenshots, if applicable, as well as details about compensation (if any)? 
    \item[] Answer: \answerNA{} 
    \item[] Justification: \answerNA{}

\item {\bf Institutional Review Board (IRB) Approvals or Equivalent for Research with Human Subjects}
    \item[] Question: Does the paper describe potential risks incurred by study participants, whether such risks were disclosed to the subjects, and whether Institutional Review Board (IRB) approvals (or an equivalent approval/review based on the requirements of your country or institution) were obtained?
    \item[] Answer: \answerNA{} 
    \item[] Justification: \answerNA{}

\end{enumerate}

\end{document}